\documentclass[10pt,twocolumn,letterpaper]{article}

\usepackage[pagenumbers]{cvpr} 

\usepackage{times}
\usepackage{epsfig}
\usepackage{graphicx}
\usepackage{amsmath}
\usepackage{amssymb}
\usepackage{booktabs}
\usepackage[ruled,vlined]{algorithm2e}
\usepackage{tabularx}
\usepackage{multirow}
\usepackage{array}
\usepackage{float}
\usepackage{makecell}
\usepackage{bbding}
\usepackage{enumitem}
\usepackage{pifont}
\usepackage[usenames,dvipsnames]{color}
\usepackage{colortbl}
\definecolor{eventcolor}{rgb}{.855, 0.890, 0.953}
\definecolor{rgbcolor}{rgb}{.886,.941,.851}
\usepackage{subcaption}
\usepackage{animate}
\usepackage{svg}

\definecolor{cvprblue}{rgb}{0.21,0.49,0.74}
\usepackage[pagebackref,breaklinks,colorlinks,citecolor=cvprblue]{hyperref}

\usepackage[capitalize]{cleveref}
\crefname{section}{Sec.}{Secs.}
\Crefname{section}{Section}{Sections}
\Crefname{table}{Table}{Tables}
\crefname{table}{Tab.}{Tabs.}

\def\loss{{\mathcal{L}}}
\def\methodname{EvRGBHand}

\def\degradation{EvRGBDegrader}

\def\multimodal{EvImHandNet}

\def\paperID{2429} 
\def\confName{CVPR}
\def\confYear{2024}

\title{Complementing Event Streams and RGB Frames for Hand Mesh Reconstruction}

\author{Jianping Jiang${}^{* 1,2,3}$, Xinyu Zhou${}^{* 4}$, Bingxuan Wang${}^{1,2,3}$, Xiaoming Deng${}^{\# 5,6}$, Chao Xu${}^{4}$, Boxin Shi${}^{\# 1,2,3}$\\
$^{1}${\normalsize National Key Laboratory for Multimedia Information Processing, School of Computer Science, Peking University}\\
$^{2}${\normalsize National Engineering Research Center of Visual Technology, School of Computer Science, Peking University}\\
$^{3}${\normalsize AI Innovation Center, School of Computer Science, Peking University}\\
$^{4}${\normalsize National Key Lab of General AI, School of Intelligence Science and Technology, Peking University}\\
$^{5}${\normalsize Institute of Software, Chinese Academy of Sciences}  \quad
$^{6}${\normalsize University of Chinese Academy of Sciences}
}

\begin{document}
\maketitle

\let\thefootnote\relax\footnotetext{${}^{*}$ equal contribution, ${}^{\#}$ corresponding author}

\begin{abstract}
 Reliable hand mesh reconstruction (HMR) from commonly-used color and depth sensors is challenging especially under scenarios with varied illuminations and fast motions. Event camera is a highly promising alternative for its high dynamic range and dense temporal resolution properties, but it lacks salient texture appearance for hand mesh reconstruction. In this paper, we propose \methodname{} -- the first approach for 3D hand mesh reconstruction with an event camera and an RGB camera compensating for each other. 
By fusing two modalities of data across time, space, and information dimensions, \methodname{} can tackle overexposure and motion blur issues in RGB-based HMR and foreground scarcity as well as background overflow issues in event-based HMR.
 We further propose \degradation{},
 which allows our model to generalize effectively in challenging scenes, even when trained solely on standard scenes, thus reducing data acquisition costs.
 Experiments on real-world data demonstrate that \methodname{} can effectively solve the challenging issues when using either type of camera alone via retaining the merits of both, and shows the potential of generalization to outdoor scenes and another type of event camera.
For code, models, and dataset, please refer to \href{https://alanjiang98.github.io/evrgbhand.github.io/}{https://alanjiang98.github.io/evrgbhand.github.io/}.
\end{abstract}    
\section{Introduction}
\label{sec:intro}

Reliable 3D hand mesh reconstruction (HMR) is essential for
various applications in virtual reality and robotics.
Although great progress on HMR has been made for color~~\cite{fastmetro22, mobrecon22, intaghand22}, depth~\cite{handtrans20, jgr20, handvoxnet++21, deng2020weakly}, and event cameras~\cite{eventhands21, differentiable21}, HMR based on a single sensor can not achieve satisfactory performance for different scenarios. The frame-based RGB or depth imaging mechanism inevitably faces degenerated issues, such as \textbf{overexposure} under strong light conditions and \textbf{motion blur} when hands move fast, which poses challenges to conducting robust HMR.

\begin{figure}[t]
  \centering
  \includegraphics[width=\linewidth]{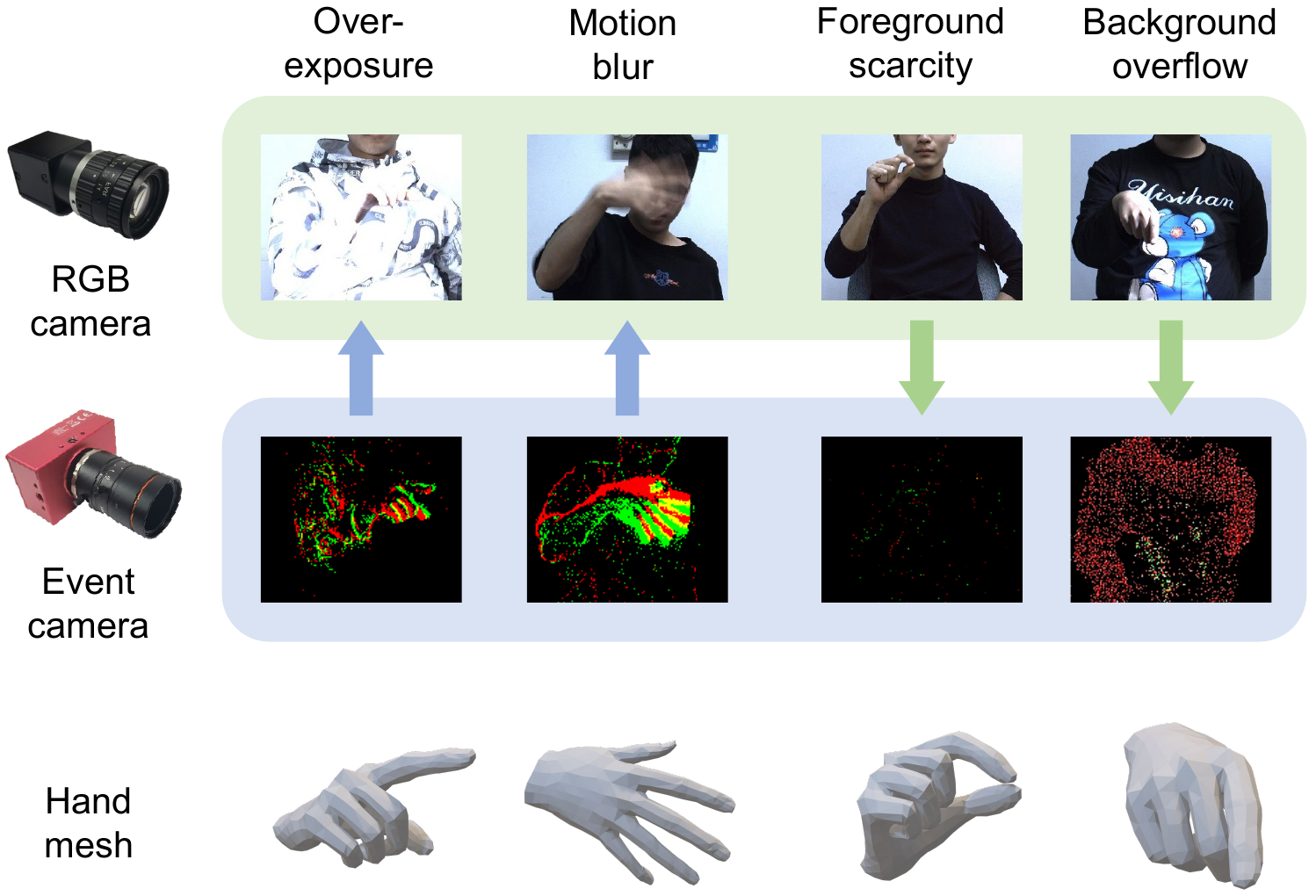}
  \caption{
   Due to the differences in RGB camera and event camera imaging mechanisms, it is promising to make complementary use of both modalities of data to achieve robust hand mesh reconstruction and tackle their respective challenging issues listed at the top.
   The arrows between the first and second rows point to the compensated data domain using the data from their tails.
  }
  \label{fig:teaser}
\end{figure}

Recently, event cameras have shown great potential in HMR for high dynamic range (HDR) and fast motion scenes~\cite{eventhands21} thanks to their superior properties from neuromorphic imaging mechanism in dynamic range and temporal resolution.
Being generated asynchronously by measuring per-pixel intensity changes, event streams~\cite{davis08} are incapable in preserving effective texture and edge information in two typical scenarios. First, events triggered from hands are rare when hands keep static (we call it ``\textbf{foreground scarcity}" issue). Second, events triggered from the background are excessive when illumination significantly changes, which can heavily confuse the events from hand motion (we call it ``\textbf{background overflow}" issue). 
We show these issues in \cref{fig:teaser}, which motivate us to combine RGB frames and event streams to compensate for each other and improve the performance on their respective issues.
Advantages from such fusion have been demonstrated on several vision tasks, such as feature tracking~\cite{Messikommer23feature}, super-resolution~\cite{evintsr21, Lu23implicit, jointfilter20}, and data association~\cite{dataasso22}, but there has been no work specially designed for HMR yet.

Combining RGB frames and asynchronous event streams for HMR faces two challenges.
First, event streams and RGB images differ in data format, space, temporal distribution, and visual information carried.
It is still an open problem to conduct a fully adaptive multi-modal fusion strategy for HMR with images and events.
Second, it is difficult to obtain high-quality 3D hand annotations, especially in challenging scenes (\eg strong light, fast motion, flash at a large scale).
Hence, \textit{how to enable models to generalize well from limited training data in normal scenes to real-world challenging scenes} remains an open problem.

To tackle these challenges, we propose \methodname{} -- an transformer-based~\cite{transformer17} framework for 3D HMR to make complementary benefits of event streams and RGB frames as shown in \cref{fig:pipeline}.
We design \multimodal{} to bridge the gap in data distribution across two modalities by spatial alignment, complementary fusion, and temporal attention on event streams and RGB images.
To effectively enhance the model's generalization capability, we further propose \degradation{}, a data augmentation module for event and image pairs, enabling our model to be trained solely on normal scenes and yet significantly improve performance in challenging settings.
To evaluate our method, we collect a real-world event-based hand dataset {\sc EvRealHands} with 3D annotations and build a large-scale synthetic dataset to enlarge training data.
Experiments on real-world data show that \methodname{} can effectively tackle the challenging issues by compensating for each other and well balance between computational cost and accuracy, even with the most vanilla transformer-based fusion strategy~\cite{perceiver21, VATT21}.
Preliminary qualitative analysis shows that \methodname{}, once trained solely on indoor scenes captured by the DAVIS346 event camera~\cite{davis08}, demonstrates cross-environment generalization to outdoor scenes and cross-camera adaptability to another type of event camera.
The main contributions of this paper can be summarized as follows:
\begin{enumerate}[noitemsep]
    \item We investigate the feasibility of using events and images for HMR, and propose the first solution to 3D HMR by complementing event streams and RGB frames.
    \item We introduce \multimodal{}, a novel approach for effectively fusing event streams and RGB images across spatial, temporal, and informational dimensions.
    \item We propose \degradation{}, a data augmentation method specifically designed for enhancing the generalization capability of models in challenging scenes for HMR with events and images.
\end{enumerate}

\section{Related work}
\vspace{-1mm}
\subsection{RGB-based HMR}
Prior works on 3D HMR can be divided into two categories: parametric and non-parametric methods~\cite{fastmetro22}. Parametric methods~\cite{3dhand19, weakly3D18, endtoend19, modelbased21} estimate the parameters of a hand model such as MANO~\cite{mano17} while non-parametric methods~\cite{weaklymesh20, mobrecon22, metro21, zhou20mocap} directly regress the positions of the hand mesh vertices. Although parametric methods involve the hand shape prior into the approaches, they ignore spatial correlations~\cite{mobilehand20} and regressing 3D rotations is a challenging task~\cite{rotation18}.
Recent network architectures such as graph convolutional neural network (GCN)~\cite{gcn17} and transformer~\cite{transformer17} significantly improve the performance of non-parametric methods.
GCN-based methods~\cite{weaklymesh20, mobrecon22} can model the vertex-to-vertex correlations, and transformer-based methods~\cite{metro21, meshgraphormer21, fastmetro22} can learn the relationships among joints and mesh vertices, thus tackling the partial occlusion issue effectively.
Considerable progress has been made in HMR based on a single RGB frame, but sequence-based studies are still inadequate. Prior sequence-based methods involve the temporal information by recurrent networks~\cite{seqhands20, multiviewvideo21} or a tracking framework~\cite{rgb2hands20, megatrack20}. However, these sequence-based methods cannot simultaneously achieve multi-modal fusion.

\begin{figure*}[t]
  \centering
  \includegraphics[width=\textwidth]{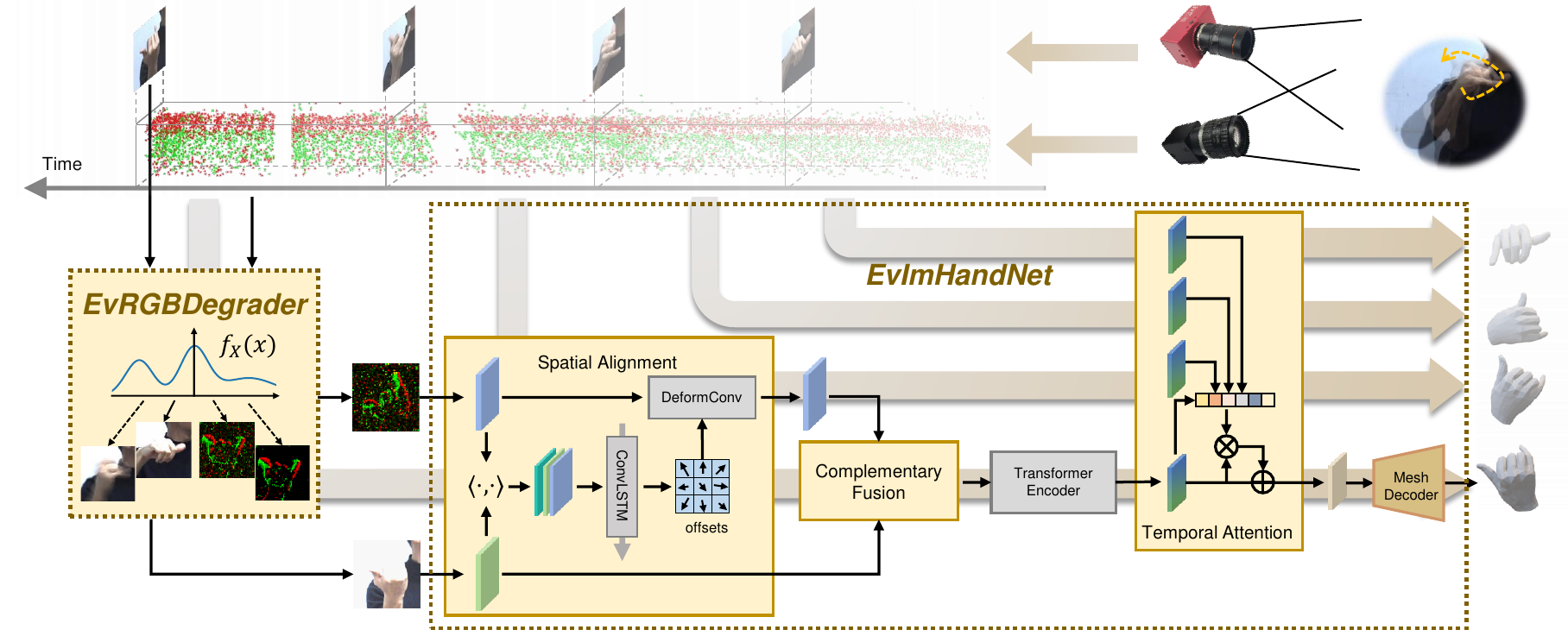}
  \caption{
  Overview of our pipeline.
  During training, we first generate various challenging scene data from normal scene sequences via \degradation{}. Then we achieve spatial alignment of the event and image features using the Deformable module with temporal motion clues. Once aligned, we feed these features subsequently to complementary fusion module (detailed architecture in \cref{fig: fusion module}) for scene-aware fusion, the transformer encoder to learn non-local correlations and mapping them to the latent hand space. We then apply temporal attention on context hand features to leverage the spatial-temporal consistency of hand motions.
  Finally, the mesh decoder maps the hand features into the 3D coordinates of hand vertices and joints. In evaluation, we deactivate \degradation{}.
  }
  \label{fig:pipeline}
\end{figure*}

\subsection{Event-based HMR}
Event cameras~\cite{davis08} generate asynchronous events by measuring per-pixel brightness changes and have several merits over RGB cameras, such as high dynamic range (120 dB), high time resolution (up to 1 $\mu$s), low redundancy, and low power consumption.
Recent researches have shown their potential in several vision tasks, such as detection~\cite{detection18}, tracking~\cite{asyntracking18}, optical flow estimation~\cite{evflow19}, super-resolution~\cite{evintsr21}, human pose estimation~\cite{eventhpe21}, \etc.
EventHands~\cite{eventhands21} is the first learning-based approach to conduct event-based 3D HMR solution and qualitatively demonstrates the benefits of event cameras for 3D HMR in strong light and fast motion scenes. 
Jalees \etal~\cite{differentiable21} propose an event-based hand tracking system in an energy-based optimization paradigm.
Since both methods are solely based on event streams, they inevitably face low spatial resolution, foreground scarcity, and background overflow issues.
As far as we know, there is no existing hand mesh reconstruction approach using both event streams and RGB frames. 
The closest work is EventCap~\cite{eventcap20}, which applies to human pose estimation from event streams and gray-scale images for the first time.
It first obtains an initial pose from gray-scale images and reconstructs human motion with high frame rate by event trajectories.
Nevertheless, the initialization from gray-scale images is not robust to strong light scenes and the fitting approach using event trajectories cannot involve the appearance information from gray-scale images.
In contrast to loose data association in EventCap~\cite{eventcap20}, our approach utilizes tight feature-level fusion of the two modalities, enabling the two cameras to complement each other in HMR.

\subsection{Event-image Fusion}
The fusion of event streams and RGB images faces diverse challenges across data format, space, time, and information dimensions. 
Current fusion approaches can be broadly categorized into two main types: pixel-level and feature-level approaches. Pixel-level approaches~\cite{tulyakov22timelens++, messikommer22multi-bracket, jointfilter20, sun22event-based, zhang23alignment} align events and images at the pixel level, leveraging the imaging constraints of event cameras for fusion. They are commonly used in low-level vision tasks. 
Feature-level methods~\cite{Lu23implicit, Messikommer23feature, tomy22fusing} align events and images in the feature space, utilizing spatial-temporal relationships for fusion, and are frequently used in middle-level and high-level vision tasks.
Since HMR aims to estimate the motion of a 3D non-rigid mesh, it is necessary to consider the complementary usage of two modal information in imaging, the spatial alignment of two free-viewpoint data, and the spatial-temporal consistency of the hand motion. 
This presents greater challenges than previous tasks.


\section{Method}
\label{sec: method}
The pipeline of \methodname{} is illustrated in \cref{fig:pipeline}.
\methodname{} consists of \multimodal{} to complement events and images for robust HMR in \cref{sec: multimodal} and \degradation{} to enable the model to generalize well in challenging scenes in \cref{sec: degradation}.
In \multimodal{}, we adopt spatial alignment, complementary fusion, and temporal attention to estimate hand shapes and 3D joints from the events and image pair.
To address the difficulty in obtaining challenging scene data with 3D annotations, we apply data augmentation on normal scene training data through \degradation{}. This effectively enhances the generalization performance of our model under challenging scenarios to outdoor scenes and another type of event camera.


\subsection{Preliminaries}
\label{sec:prelim}
\noindent \textbf{Hand model representation.}
We adopt a differentiable hand parametric MANO model \cite{mano17} as hand model representation. Mesh vertices of MANO can be obtained by function $\mathbf{V} = M(\boldsymbol{\theta}, \boldsymbol{\beta}) \in \mathbb{R}^{778 \times 3}$ and 3D joints $\mathbf{J}_{\text{3D}} \in \mathbb{R}^{21 \times 3}$ can be recovered by regression function $\mathbf{J}_{\text{3D}} = J_{\text{reg}}(M({\boldsymbol{\theta}}, {\boldsymbol{\beta}}))$ with pose parameters $\boldsymbol{\theta}$ and shape parameters $\boldsymbol{\beta}$.

\noindent \textbf{Event camera.}
Event cameras generate asynchronous event streams by recording the changes of per-pixel intensity $I(x, y, t)$. An event $e_i = (x_i, y_i, t_i, p_i)$ is triggered at pixel $(x_i, y_i)$ at time $t_i$ when the logarithmic brightness change meets the condition:
\begin{equation}
    \log I(x_i, y_i, t_i) - \log I(x_i, y_i, t_p) = p_i C,
\end{equation}
where $t_p$ is the last event triggering timestamp at the same pixel, $p_i \in \{ -1, 1\}$ is the polarity, $C$ is the threshold.

\subsection{\multimodal{}}
\label{sec: multimodal}

To make the asynchronous event streams compatible with modern deep learning architectures~\cite{hrnet21, transformer17}, we use the time surface representation from EventHands~\cite{eventhands21}.
Considering events triggered from the hand at timestamp $t$ are sparse, we use $N$ events (denoted as $E_{t}^{N}$) before timestamp $t$ to form a two-channel stacked frame $I_{\text{Ev}, t}$ by iterating each event $e_i$ in $E_{t}^{N}$ as:
\begin{equation}
    I_{\text{Ev}, t}(x_i, y_i, p_i) = \frac{t_i - t_{s} }{t - t_{s}},
\end{equation}
where $t_s$ is the timestamp of the first event in $E_{t}^{N}$.
The stacked frame $I_{\text{Ev}, t}$ with two channels can effectively record hand motions by assigning higher weights to events closer to the target time.

\paragraph{Spatial alignment.}
Since HMR is a task that estimates the 3D coordinates of hand vertices and joints from camera observations, aligning spatial information in both events and images is crucial. 
In practical applications, events and images can be captured from the same viewpoint, such as in DAVIS~\cite{davis08}, or from different viewpoints, as seen in hybrid cameras~\cite{tulyakov22timelens++, evflow19}. 
Consequently, the approach based on epipolar geometry~\cite{he2020epipolar, yang2023poem} lacks generality. 
Meanwhile, methods based on cost volumes~\cite{yang2020cost} or vanilla transformer architectures~\cite{meshgraphormer21} have a high computational cost, which is not suitable with the low-power nature of event camera.
To address these challenges, we directly achieve HMR through the correlation between the data, being unaware of the relative camera positions.

To achieve spatial alignment between two modalities, we first use a shallow CNN module $f^{\text{C}}$ (ResNet34~\cite{resnet16}) to extract 24$\times$24 feature maps $F_{\text{Im}, t}^{\text{C}}$, $F_{\text{Em}, t}^{\text{C}}$ from the images $I_{\text{Im}, t}$ and event stacked frames $I_{\text{Ev}, t}$.
Further, drawing inspiration from the Deformable Convolution~\cite{dai17deformable, weng2018deformable}, we use the feature maps to estimate the offsets between events and images. To alleviate the temporal jitter in spatial alignment caused by texture mismatching between events and images, we exploit the temporal motion clues via a ConvLSTM\cite{Shi2015ConvolutionalLN} layer:
\begin{equation}
    \Delta P = \text{ConvLSTM}(F_{\text{Im}, t}^{\text{C}}, F_{\text{Ev}, t}^{\text{C}}),
\end{equation}
where $\Delta P$ are the offsets.
Since the offsets are learned from the feature correlation between events and images, we can achieve alignment without estimating the relative camera pose or disparity. 
Leveraging these offsets, we can obtain the aligned features $F_{t}^{\text{A}}$ of events and images using Deformable Convolution $f^{\text{DC}}$~\cite{dai17deformable}:
\begin{equation}
    F_{\text{Ev}, t}^{\text{A}} = f^{\text{DC}}(F_{\text{Ev}, t}^{\text{C}}, \Delta P).
\end{equation}

\begin{figure}
    \centering
    \includegraphics[width=\linewidth]{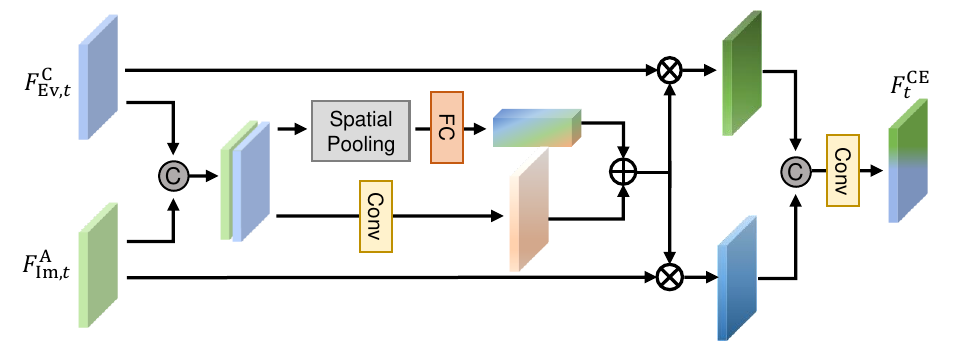}
    \caption{Detailed architecture of complementary fusion module.}
    \label{fig: fusion module}
\end{figure}

\paragraph{Complementary fusion.}
Given the complementary nature of events and images, we expect our model to learn the relationship between scene and feature selection for robust HMR. To this end, we design the complementary fusion module $f^{\text{CF}}$ \cite{woo2018cbam, yang2023learning} as illustrated in \cref{fig: fusion module}, which can automatically compute weights based on the two modality features to obtain the complementary features:
\begin{equation}
    F_{t}^{\text{CF}} = f^{\text{CF}}(F_{\text{Ev}, t}^{\text{A}}, F_{\text{Im}, t}^{\text{C}}),
\end{equation}
where $F_{t}^{\text{CF}}$ are down-sampled to 8$\times$8 for latter processing.

Inspired by FastMETRO~\cite{fastmetro22}, we use the transformer encoder framework to build non-local relationships among the complementary features.
The features $F_{t}^{\text{CF}}$ are flattened as transformer tokens, and fed into the transformer encoder $f^{\text{TE}}$ which consists of $L$ sequential transformer blocks. The outputs of transformer blocks are latent hand features $F_{t}^{\text{H}} = \{F_{t}^{l}, l=1,2,...,L\}$:
\begin{equation}
    F_{t}^{\text{H}} = f^{\text{TE}}(F_{t}^{\text{CF}}).
\end{equation}
The transformer encoder can effectively exploits non-local associations of hand observations within the feature map, addressing the self-occlusion issue in HMR.

\paragraph{Temporal attention.}
Hand motion exhibits spatial-temporal continuity, and the event streams contain rich temporal and motion information. Therefore, we propose a temporal attention mechanism to effectively leverage the hand motion context information. 
We employ relative position encoding~\cite{shaw18relative} to apply temporal attention $f^{\text{TA}}$ for each token within the hand feature for sequential $S$ steps:
\begin{equation}
    F_{t}^{\text{TAH}}(x, y) = f^{\text{TA}}(\{F_{t+s}^{\text{H}}(x, y), \quad s=-S, ..., 0\}),
\end{equation}
where $F_{t}^{\text{TAH}}$ are the final latent hand features.
On one hand, the temporal attention mechanism ensures smooth hand mocap. On the other hand, it can utilize motion information from other moments to compensate for the current instance, leading to more stable HMR.

We use a transformer decoder architecture with $L$ transformer blocks to regress the mesh vertices and joints, which has also been adopted in FastMETRO~\cite{fastmetro22}.
The transformer decoder takes the learnable joint tokens $\{\mathbf{q}_1^{J}, \mathbf{q}_2^{J}, ..., \mathbf{q}_{21}^{J}\}$ and vertex tokens $\{\mathbf{q}_1^{V}, \mathbf{q}_2^{V}, ..., \mathbf{q}_{195}^{V}\}$ as input, where $\mathbf{q}_i^{J}, \mathbf{q}_i^{V} \in \mathbb{R}^{D}$.
Given latent hand features $F_{t}^{\text{TAH}}$, the transformer decoder learns non-local correlations among vertices and joints by passing joint and vertex features through cross-attention and self-attention layers.
An MLP-based 3D coordinate regressor estimates the hand mesh vertices of the coarse mesh and 3D joints using the outputs of the transformer decoder.
For mesh vertices, we use an MLP layer to upsample the coarse mesh (195 vertices) to a fine mesh (778 vertices) as the hand MANO model.

\begin{figure}[t]
  \centering
  \subfloat[Color jitter augmentation for overexposure]{
     \centering
     \includegraphics[width=\linewidth]{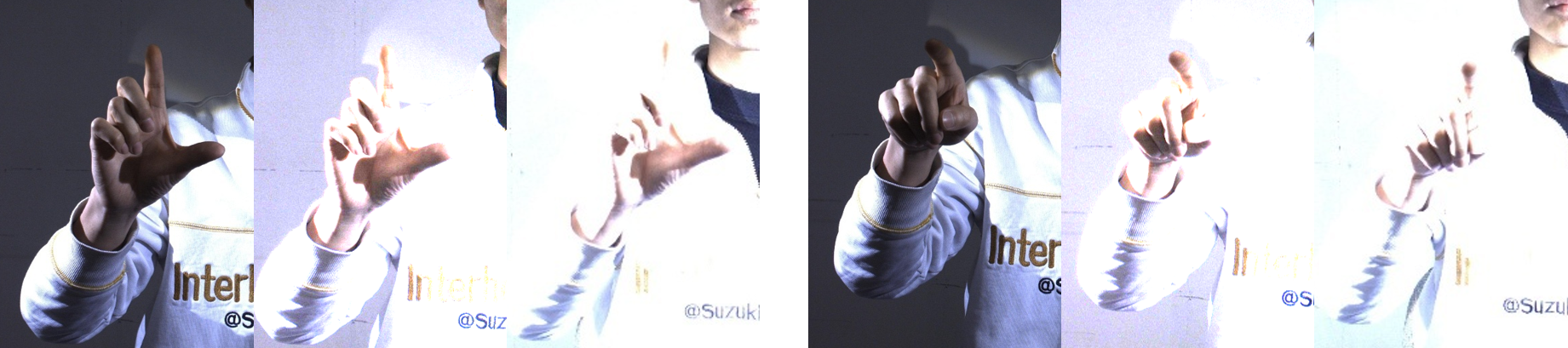}
  }\\
      \subfloat[Salt and pepper noise for background overflow]{
     \centering
     \includegraphics[width=\linewidth]{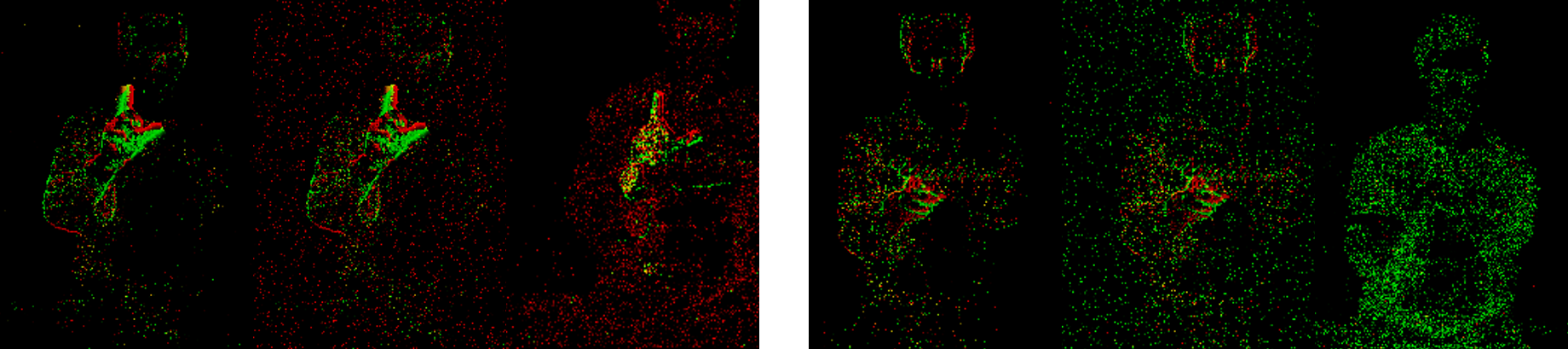}
  }\\
  \caption{
    Visualization of train-evaluation gap and \degradation{}.
    For each triplet from left to right, we show original data, degraded data, real data with challenging issues.
  }
  \label{fig: augmentation}
\end{figure}

\subsection{\degradation{}}
\label{sec: degradation}

\begin{figure}[t]
    \centering
    \includegraphics[width=\linewidth]{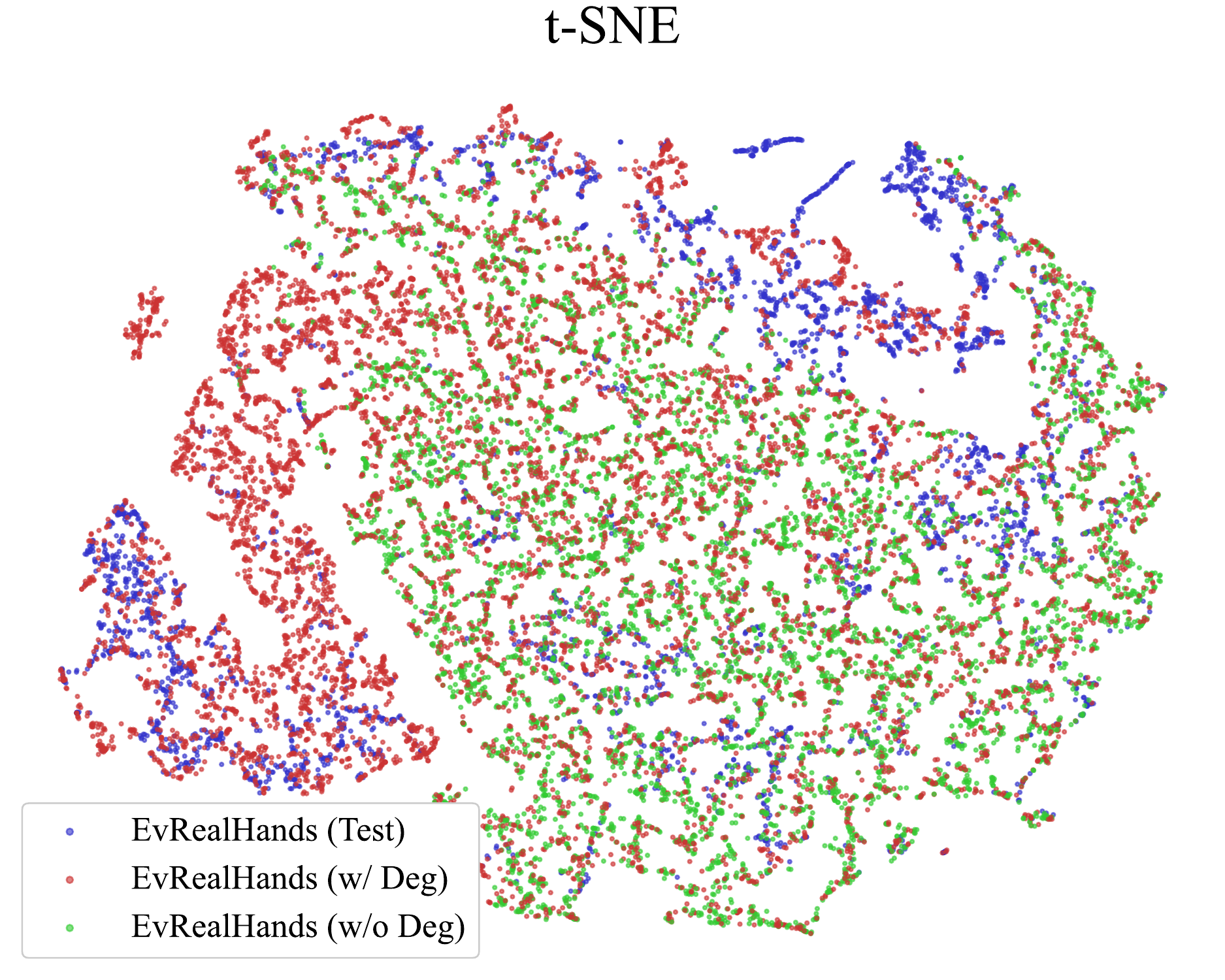}
    \caption{Visualization for events and image descriptor vectors by t-SNE. The descriptor vector has four dimensions: image sharpness, image brightness, and the means of positive and negative polarity events.}
    \label{fig: degradation}
\end{figure}

The acquisition and annotation of high-quality 3D hand datasets are of high cost, especially under challenging scenes such as strong light, fast motion, and flash. 
This prompts us to leverage data under normal scenes to endow models with the capability to generalize to challenging scenes.
As shown in \cref{fig: augmentation}, we observe that for the data pair $(I_{\text{Im}}, I_{\text{Ev}})$, the degradation process under challenging conditions is traceable. For instance, the brightness of image $I_{\text{Im}}$ is high under strong light, while the distribution of $I_{\text{Ev}}$ remains relatively stable. 
In flashing scenes, the mean value of $I_{\text{Ev}}$ increases significantly along a dimension, while the texture and sharpness of $I_{\text{Im}}$ are little affected. 
Therefore, \degradation{} consists of three core augmentations:
\begin{itemize}
    \item \textbf{Overexposure (OE): }For RGB frames, we use color jitter augmentation to change the image brightness and augment the strong light scenes.
    \item \textbf{Motion blur (MB): }For simulating motion blur, we warp the original image with optical flow following~\cite{farneback03flow} in OpenCV to interpolate frames and average them.
    \item \textbf{Background overflow (BO):} We add salt and pepper noise on training event stacked streams to simulate the leak noise. Each pixel of the stacked frames will emit salt and pepper noise randomly.
\end{itemize}
During the training process, we apply degradation to a data pair $(I_{\text{Im}}, I_{\text{Ev}})$ at a certain probability to yield a degraded data pair $(I^{\text{DG}}_{\text{Im}}, I^{\text{DG}}_{\text{Ev}})$:
\begin{equation}
    (I^{\text{DG}}_{\text{Im}}, I^{\text{DG}}_{\text{Ev}}) = f_{X}((I_{\text{Im}}, I_{\text{Ev}})),
\end{equation}
where $f_{X}$ is the degradation probability distribution of OE, MB, and BO.
The t-SNE visualization in \cref{fig: degradation} implies that challenging scenes (test) and normal scenes (w/o Deg) exhibit distribution gap in the imaging descriptor space, which can be bridged by \degradation{} (w/ Deg)

\subsection{Training}
\label{sec:training}
Following the common practice in transformer-based mesh reconstruction method~\cite{meshgraphormer21, metro21, fastmetro22}, we use vertex loss and joint loss as supervisions on each predicted result:
\begin{eqnarray}
    &&\loss_{\mathbf{V}}=\frac{1}{M}{\lVert \mathbf{V} - \hat{\mathbf{V}} \rVert_{1}}, \quad \loss_{\mathbf{J}}=\frac{1}{K}{\lVert \mathbf{J} - \hat{\mathbf{J}} \rVert_{2}},
\end{eqnarray}
where ${\mathbf{V}}, {\mathbf{J}}$ are predicted mesh vertices and 3D joints, $\hat{\mathbf{V}}, \hat{\mathbf{J}}$ are respective ground truths, $M=778$, and $K=21$.


For supervision on sequential data, the total loss is the sum of vertex losses and joint losses of one hand mesh from RGB-based HMR and $S$ sequential hand meshes from event-based HMR:
\begin{align}
    \loss_{\text{all}} = &\lambda_{\mathbf{V}}\loss_{\mathbf{V}}^{{\text{Im}}} + \lambda_{\mathbf{J}}\loss_{\mathbf{J}}^{{\text{Im}}} + \sum_{s=1}^{S}(\lambda_{\mathbf{V}}\loss_{\mathbf{V}, s}^{{\text{Ev}}} + \lambda_{\mathbf{J}}\loss_{\mathbf{J}, s}^{{\text{Ev}}}).
\end{align}

\section{Datasets and metrics}
To demonstrate our method under various challenging scenarios, we collect the real-world event-based hand dataset {\sc EvRealHands} with 3D annotations, which covers the typical challenging issues for RGB images and events (examples in \cref{fig:teaser}).
To supplement training data for better performance,
we develop a synthetic dataset from the RGB-based hand dataset {\sc InterHand2.6M}~\cite{interhand20}. 

\begin{table}[t]
    \centering
    \caption{Scenes and their corresponding issues that challenge RGB or event-based HMR in our {\sc EvRealHands} datasets.  
    (FG and BG are short for foreground and background.)}
    \resizebox{\linewidth}{!}{
    \begin{tabular}{ccccc}
    \toprule
    \multirow{3}{*}[-9pt]{Scenes} &  \multicolumn{2}{c}{RGB} & \multicolumn{2}{c}{Event} \\
    & \makecell[c]{Overexposure} & \makecell[c]{Motion blur} & \makecell[c]{FG scarcity}  & \makecell[c]{BG overflow} \\
     &  
    \begin{minipage}[b]{0.3\columnwidth}
		\centering
		\raisebox{-.5\height}{\includegraphics[width=\linewidth]{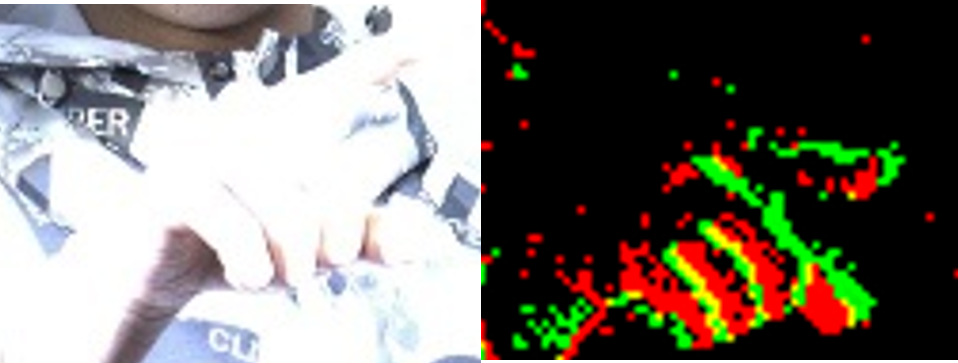}}
	\end{minipage}
         & 
        \begin{minipage}[b]{0.3\columnwidth}
		\centering
		\raisebox{-.5\height}{\includegraphics[width=\linewidth]{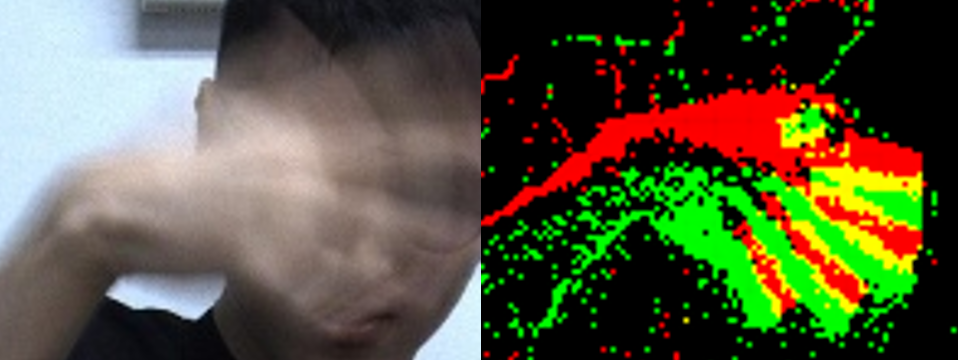}}
	\end{minipage} & 
	\begin{minipage}[b]{0.3\columnwidth}
		\centering
		\raisebox{-.5\height}{\includegraphics[width=\linewidth]{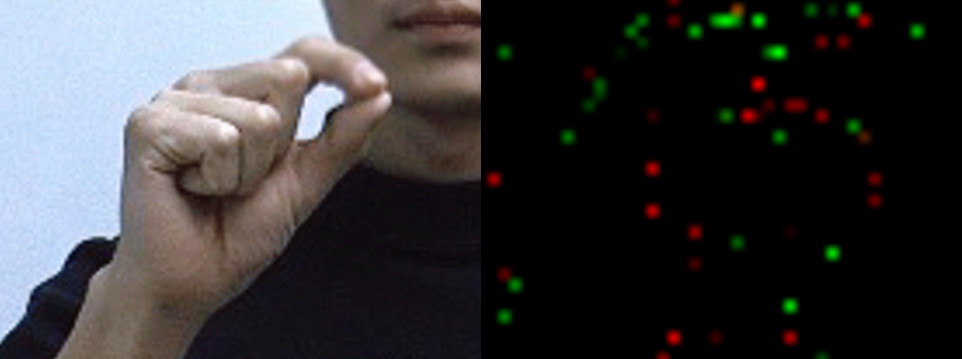}}
	\end{minipage}  & 
	\begin{minipage}[b]{0.3\columnwidth}
		\centering
		\raisebox{-.5\height}{\includegraphics[width=\linewidth]{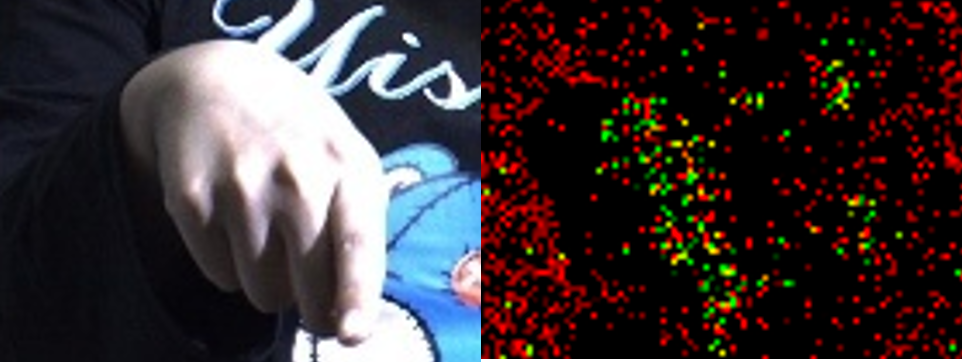}}
	\end{minipage}\\
    \cmidrule(r){1-5}
     Normal & \textbf{—} & \textbf{—}  & {\color{black}\Checkmark}& \textbf{—} \\
     Strong light & {\color{black}\Checkmark} & \textbf{—} &{\color{black}\Checkmark} & \textbf{—} \\
    Flash &\textbf{—} & \textbf{—} &{\color{black}\Checkmark} & {\color{black}\Checkmark} \\
     Fast motion & \textbf{—} & {\color{black}\Checkmark} & \textbf{—}&  \textbf{—}\\
    \bottomrule
    \end{tabular}}
    \label{table: scene issue corr}
\end{table}

\subsection{Real-world data}
The indoor sequences of {\sc EvRealHands} are captured using a multi-camera system following~\cite{handdataset17, megatrack20} with 7 RGB cameras (FLIR, 2660$\times$2300 pixels, 15 FPS) and an event camera (DAVIS346, 346$\times$260 pixels) capturing data from different views simultaneously.
We collect 4,452 seconds of event streams and RGB images from 10 subjects. Each subject performs 15 fixed poses~\cite{ges2017} and random hand poses. To include challenging issues caused by RGB and event imaging mechanisms, we set up strong light, flash, and fast motion scenes in addition to normal scenes. The scenes and their corresponding issues are listed in \cref{table: scene issue corr}.
Additionally, we capture data in outdoor scenes through a hybrid camera system for qualitative evaluation.
The system is composed of an RGB camera (FLIR BFS-U3-51S5) and an event camera (DAVIS346 Mono~\cite{davis08} or PROPHESEE GEN 4.0~\cite{thomas2020prophesee}) via a beamsplitter (Thorlabs CCM1-BS013).
We collect 12 outdoor sequences (6 for DAVIS346, 6 for PROPHESEE) from 3 subjects, including sequences with fast motion, variant illuminations.

\begin{figure*}[t]
  \centering
  \includegraphics[width=\textwidth]{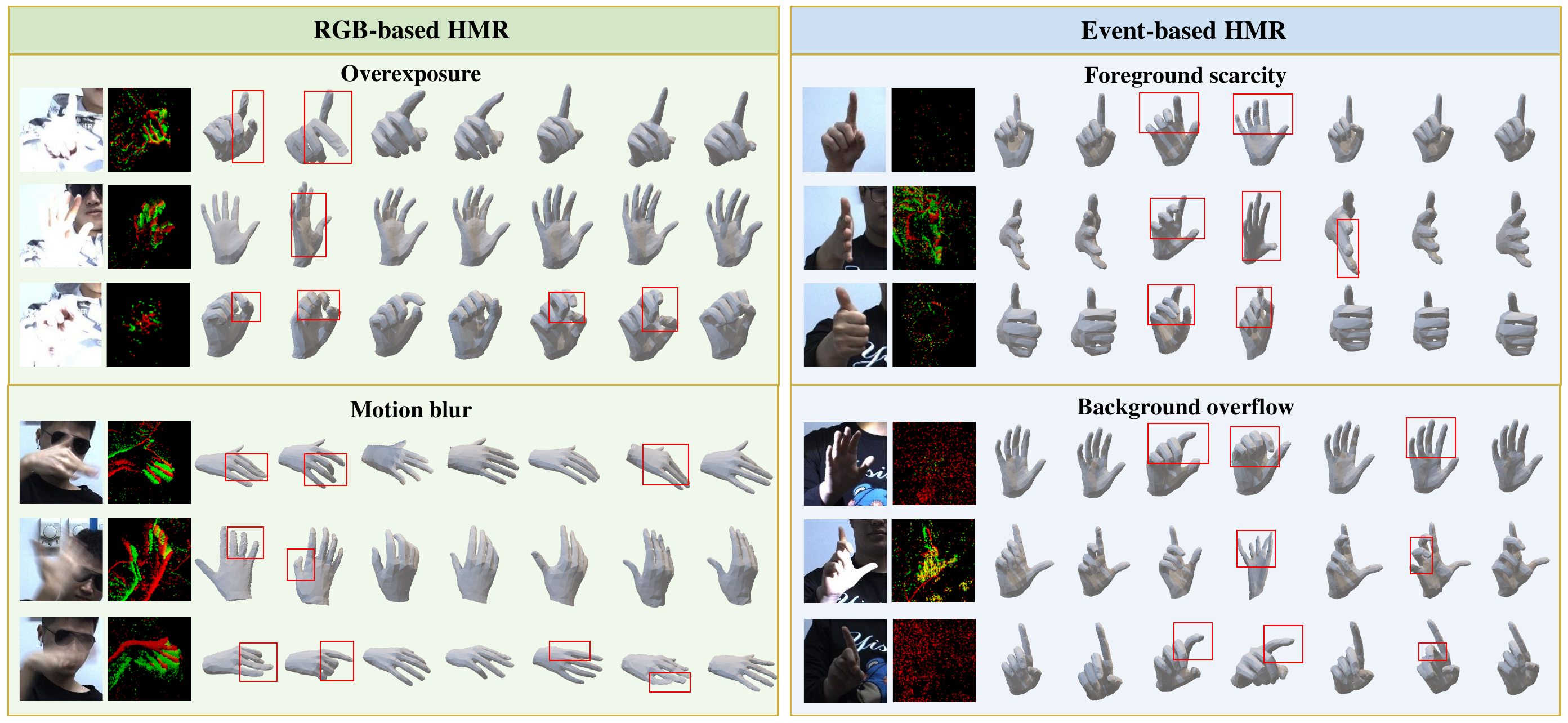}
  \vspace{-0.7cm}
  \begin{flushleft}
    	\scriptsize  \hspace{0.4cm}RGB~ \hspace{0.25cm} Events~ \hspace{0.1cm} MG~\cite{meshgraphormer21} \hspace{0.02cm} FR~\cite{fastmetro22} \hspace{0.05cm} EH~\cite{eventhands21}  \hspace{0.13cm}~ FE\hspace{0.1cm}\quad~ Vanilla \hspace{0.15cm} w/o Deg \hspace{0.15cm} Ours  \qquad~ RGB~ \hspace{0.15cm} Events~ \hspace{0.1cm} MG~\cite{meshgraphormer21} \hspace{0.03cm} FR~\cite{fastmetro22} \hspace{0.05cm} EH~\cite{eventhands21}  \hspace{0.13cm}~ FE\hspace{0.1cm}\quad~ Vanilla \hspace{0.15cm} w/o Deg \hspace{0.1cm} Ours
  \end{flushleft}

  \caption{
    Qualitative analysis of HMR methods under challenging issues. For each issue, columns from left to right are RGB images, events, results from Mesh Graphormer (MG)~\cite{meshgraphormer21}, FastMETRO-RGB (FR)~\cite{fastmetro22}, EventHands (EH)~\cite{eventhands21}, FastMETRO-Event (FE), \methodname{}-vanilla (Vanilla), \methodname{} without \degradation{} (w/o Deg) and \methodname{} (Ours).
    For easy reference, results of issues from RGB images (left side) are aligned to the event camera view and results of issues from events (right side) are aligned to the RGB camera view. 
    \methodname{} successfully tackles challenging issues of RGB images and event streams by compensating for each other.}
  \label{fig:quality results}
\end{figure*}

\subsection{Synthetic data}
To better model the distribution of real hand poses, we synthesize event streams from existing real RGB datasets.
We apply v2e~\cite{v2e21} event simulator on {\sc InterHand2.6M}~\cite{interhand20} to get synthetic event streams from RGB sequences.
We select right hand sequences of 9 camera views from 4 subjects for simulation.


\section{Experiments}

In this section, we first introduce the experimental settings in \cref{sec: settings}. We then show the experimental results of \methodname{} to demonstrate the complementary effects, generalization, and efficiency in \cref{sec: results}. We also show the ablation studies in \cref{sec: ablation}.
More information about the dataset, experimental results can be found in our video and supplementary material.

\subsection{Settings}
\label{sec: settings}


\noindent \textbf{Baselines.}
In order to demonstrate the complementary benefits of events, we compare our method with Mesh Graphormer~\cite{meshgraphormer21}, and FastMETRO~\cite{fastmetro22} (denoted as FastMETRO-RGB), which are RGB-based methods on the top of FreiHand~\cite{freihand19} leaderboard.
For event-based HMR, we use EventHands~\cite{eventhands21}, the only event-based HMR method with learning framework, as one of the baselines.
Considering that EventHands~\cite{eventhands21} is a parametric approach, a comparison between EventHands~\cite{eventhands21} and our non-parametric approach is not sufficient in demonstrating the complementary benefits of RGB images.
Therefore, we introduce FastMETRO-Event, which uses the same architecture as FastMETRO~\cite{fastmetro22} and takes the same stacked event frames as input.
While there are no existing methods for HMR using both events and images, we believe comparing \methodname{} with HMR based on a single sensor would be unfair. Drawing inspiration from recent advancements in the multi-modal domain~\cite{perceiver21, VATT21, parametereffi21, flava22}, we introduce a vanilla version of event and RGB fusion for HMR (denoted as ``\methodname{}-vanilla"). 
Built upon the FastMETRO~\cite{fastmetro22} architecture, it directly inputs the event features $F_{\text{Im}, t}^{\text{C}}$ and the image features $F_{\text{Em}, t}^{\text{C}}$ as tokens into the transformer encoder for fusion. The detailed architecture can be found in the supplementary materials.

\noindent \textbf{Training and evaluation data.}
We collect 24 sequences of normal scenes 8 subjects in {\sc EvRealHands} and all the {\sc InterHand2.6M}~\cite{interhand20} synthetic data as training data.
And we set indoor sequences from the rest 2 subjects and all the outdoor sequences in {\sc EvRealHands} as evaluation data. 
Our evaluation data include 4 sequences of normal scenes, 5 sequences under strong light, 2 sequences under flash light, and 3 sequences of fast motion.
Following \cite{meshgraphormer21, fastmetro22}, we only use the right hand data.
We conduct both quantitative and qualitative evaluations on indoor data with 3D annotations. For data without 3D annotations (fast motion or outdoor sequences), we conduct qualitative assessments.

\begin{table}[t] 
    \centering
    \caption{Quantitative comparison among HMR based on a single sensor or complementary usage in several scenes.}
    \resizebox{\linewidth}{!}{
    \begin{tabular}{ccccc}
    \toprule
    Scenes & Methods & MPJPE $\downarrow$ & MPVPE $\downarrow$ & PA-MPJPE $\downarrow$\\
    \cmidrule(r){1-5}
    \multirow{6}{*}{Normal}& Mesh Graphormer~\cite{meshgraphormer21} &  11.57 & 11.68 &  5.49\\
    & FastMETRO-RGB~\cite{fastmetro22} & 11.71 & 12.03 & 5.56 \\
    \cmidrule(r){2-5}
    & EventHands~\cite{eventhands21} & 21.13 & 20.12 & 9.05  \\
    & FastMETRO-Event & 18.36 & 17.81 & 7.85 \\
    \cmidrule(r){2-5}
    & \methodname{}-vanilla & 11.84 & 11.98 & 5.07 \\
    & Ours & \textbf{11.47} & \textbf{11.63} & \textbf{5.02} \\
    \cmidrule(r){1-5}
    \multirow{7}{*}{\makecell[c]{Strong \\ Light}} & Mesh Graphormer~\cite{meshgraphormer21} &  40.59 & 38.19 & 13.96 \\
       & FastMETRO-RGB~\cite{fastmetro22} & 35.02 & 33.52 & 13.53\\
    \cmidrule(r){2-5}
    & EventHands~\cite{eventhands21} & 27.17 & 25.88 & 9.99 \\
    & FastMETRO-Event & 23.75 & 22.81 & 9.67\\
    \cmidrule(r){2-5}
    & \methodname{}-vanilla & 25.26 & 24.12 & 10.01\\
    & Ours & \textbf{22.34} & \textbf{21.36} & \textbf{9.47}\\

    \cmidrule(r){1-5}
    \multirow{7}{*}{Flash } & Mesh Graphormer~\cite{meshgraphormer21} & 23.41 & 22.85 & 10.09 \\
    & FastMETRO-RGB~\cite{fastmetro22} & 24.43 & 23.99 & 9.69\\
    \cmidrule(r){2-5}
    & EventHands~\cite{eventhands21} & 53.69 & 51.29 & 14.37 \\
    & FastMETRO-Event & 36.30 & 35.29 & 13.38\\
    \cmidrule(r){2-5}
    & \methodname{}-vanilla & 23.13 & 22.88 & 10.02\\
    & Ours & \textbf{20.44} & \textbf{20.47} & \textbf{8.98}\\
    \bottomrule
    \end{tabular}
    }
    \label{table: quantitative results}
\end{table}

\subsection{Results}
\label{sec: results}
\paragraph{Complementary effects on imaging issues.}
As quantitative results shown in \cref{table: quantitative results} and qualitative results shown in \cref{fig:quality results}, \methodname{} outperforms HMR methods based on a single RGB camera or event camera and the vanilla fusion method.
\methodname{} outperforms Mesh Graphormer~\cite{meshgraphormer21} and FastMETRO-RGB~\cite{fastmetro22}
on MPJPE 12 $\sim$ 18 mm lower in strong light scenes.
As shown in \cref{fig:quality results}, HMR methods based on RGB cameras face overexposure and motion blur issues under strong light and fast motion scenes.
\methodname{} can leverage the stable event sequences to compensate for these issues.
For event-based HMR, \methodname{} outperforms EventHands~\cite{eventhands21} and FastMETRO-Event on MPJPE 7 $\sim$ 33 mm lower in normal and flash scenes.
Results from \cref{fig:quality results} show that the failure of event-based methods in these scenes derives from the dynamic imaging mechanism, low texture information and noises.
However, \methodname{} can utilize the rich texture information and high pixel resolution of RGB images to improve the performance via complementary fusion.
Results of \methodname{}-vanilla and \methodname{} in \cref{table: quantitative results} and \cref{fig:quality results} indicate that, compared to the method that employs transformers for direct fusion, meticulously considering the relationships between the two modalities in spatial, temporal, and information dimensions can yield superior performance enhancements with limited training data. 
 Furthermore, the results of \methodname{}-vanilla also suggest that even with the most rudimentary fusion strategy, using events and images for HMR can achieve better performance than those methods based on a single sensor by MPJPE 2 $\sim$ 16 mm lower, underscoring the potential of HMR with events and images.

\begin{figure}[t]
  \centering
  \includegraphics[width=\linewidth]{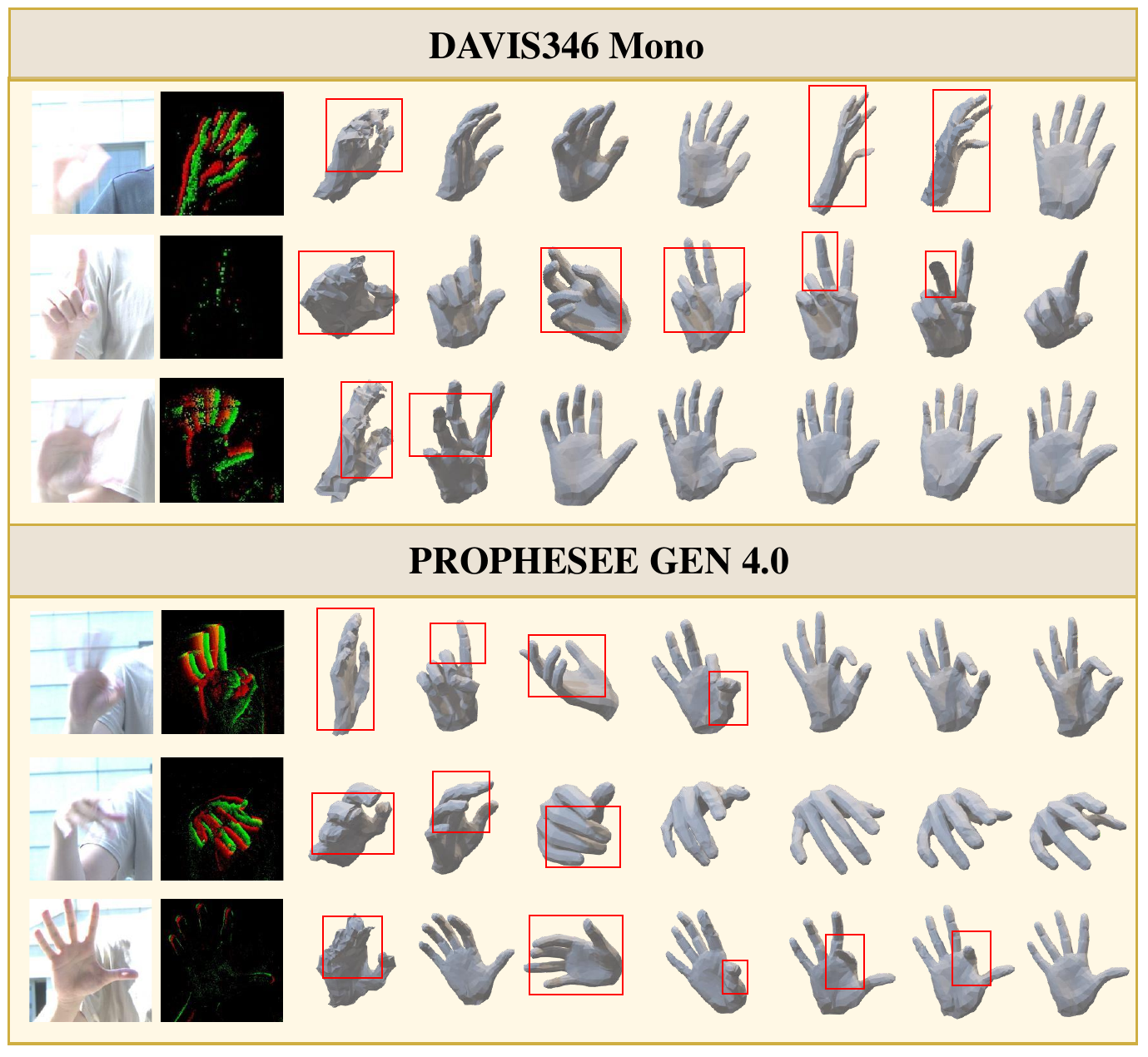}
  \vspace{-0.3cm}
  \begin{flushleft}
    	\scriptsize  \hspace{0.4cm}RGB~ \hspace{0.15cm} Events~ \hspace{0.05cm} MG~\cite{meshgraphormer21} \hspace{0.002cm} FR~\cite{fastmetro22} \hspace{0.05cm} EH~\cite{eventhands21}  \hspace{0.1cm}~ FE\hspace{0.1cm}\quad~ Vanilla \hspace{0.1cm} w/o Deg \hspace{0.1cm} Ours  
  \end{flushleft}

  \caption{
    Qualitative analysis of HMR methods on outdoor DAVIS346 sequences and PROPHESEE GEN 4.0 sequences. 
    \methodname{} generalizes better than other methods.
    }
  \label{fig:outdoor}
\end{figure}

\paragraph{Generalization.}
As qualitative results shown in \cref{fig:outdoor}, although \methodname{} was trained under normal indoor scenes using the DAVIS346 camera~\cite{davis08}, it still generalize well in challenging outdoor environments (natural various lighting, fast motion) and data captured by the PROPHESEE GEN 4.0~\cite{thomas2020prophesee}, significantly outperforming other methods. This can be attributed, on one hand, to our fusion strategy across temporal, spatial, and informational dimensions. On the other hand, it derives from the efforts of \degradation{} in bridging the distribution gap between indoor-outdoor data and normal-challenging scenes.

\paragraph{Efficiency.}
As shown in \cref{table: efficiency}, \methodname{} has 60.5\% fewer Params and requires 24.5\% fewer FLOPs than FastMETRO-Event, while achieves better performance with 6.9 mm average MPJPE lower.
Compared with \methodname{}-vanilla, \methodname{} with carefully designed architecture can achieves 79.8\% fewer Params and 54.5\% fewer FLOPs with better average accuracy.

\begin{table}[t]
    \centering
    \caption{Computational cost and average accuracy.}
    \resizebox{\linewidth}{!}{
    \begin{tabular}{ccccc}
    \toprule
    Methods & Params$\downarrow$& FLOPs$\downarrow$& MPJPE$\downarrow$& MPVPE$\downarrow$\\
     \cmidrule(r){1-5}
     EventHands~\cite{eventhands21} & 22.68 M & 2.81 G & 30.44 & 29.24 \\
    FastMETRO-Event & 141.68 M & 10.79 G & 23.59 & 23.12 \\
    \methodname{}-vanilla & 277.02 M & 17.90 G & 17.45 & 17.30 \\
     Ours & 55.92 M & 8.15 G & 16.66 & 16.43\\
    \bottomrule
    \end{tabular}}
    \label{table: efficiency}
    \vspace{-0.05cm}
\end{table}

\subsection{Ablation Studies}
\label{sec: ablation}

\paragraph{\multimodal{}.}
As quantitative results shown in \cref{sec: ablation}, spatial alignment (SA), complementary fusion (CF), and temporal attention (TA) all contribute to the stable HMR performance.
Compared to the vanilla fusion strategy, these modules collectively lead to an improvement of 2.5 $\sim$ 3 mm MPJPE in challenging scenes.

\paragraph{\degradation{}}
Quantitative results in \cref{sec: ablation} shows that the simulations of overexposure (OE) and background overflow (BO) significantly improve the performance on indoor challenging scenes (8 $\sim$ 20 mm MPJPE lower).
Qualitative results in \cref{fig:quality results} and \cref{fig:quality results} between ``w/o Deg" and ``Ours" show that \degradation{} effectively promote the performance in strong light and fast motion scenes.
This indicates that \degradation{} can effectively bridge the data distribution gap between normal collection settings and outdoor evaluation scenarios.

\begin{table}[t]
    \centering
    \caption{Ablation studies.}
    \resizebox{\linewidth}{!}{
    \begin{tabular}{ccccccccc}
    \toprule
    \multicolumn{3}{c}{\multimodal{}} & \multicolumn{3}{c}{\degradation{}}  & \multicolumn{3}{c}{ MPJPE (mm)$\downarrow$} \\
    SA & CF & TA & OE & MB & BO & Normal & Strong light & Flash \\
    \cmidrule(r){1-9}
    \ding{55} & \ding{55} & \ding{55} & & & & 11.81 & 25.35 & 22.99 \\
     & \ding{55} & \ding{55} & & & & 11.60 & 24.23 & 22.86 \\
     &  & \ding{55} & & & & 11.57 & 23.87 & 22.52 \\
     \cmidrule{1-9}
     & & & \ding{55}  & & & 11.53 & 45.11 & 28.52 \\
     &  & & & \ding{55} & & 11.48 & 23.50 & 21.13 \\
     &  &  & & & \ding{55} & 11.63 & 27.83 & 23.33 \\
    &  &  & \ding{55} & \ding{55} & \ding{55} & 11.73 & 47.34 & 29.02 \\   
       \cmidrule(r){1-9}
     &  &  & & &  & \textbf{11.47} & \textbf{22.34} & \textbf{20.43} \\

    \bottomrule
    \end{tabular}
    }
    \label{tab: ablation results}
\end{table}

\begin{figure}[t]
  \centering
  \includegraphics[width=\linewidth]{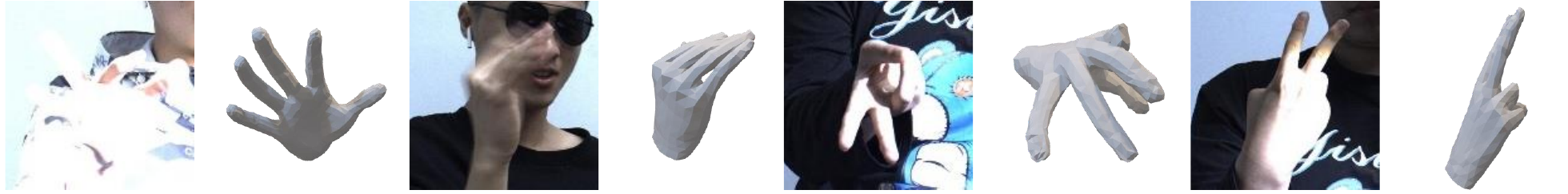}
  \caption{
   RGB images and failure cases of \methodname{}.
  }
  \label{fig:failure cases}
\end{figure}

\section{Conclusion}
In this paper, we explore the potential of complementary usage of event cameras and color cameras for hand mesh reconstruction tasks. 
To this end, we introduce a framework, \methodname{}, which leverages the strengths of both event camera and color camera imaging to achieve robust and efficient HMR. 
Through multi-modal information fusion and degradation augmentation, our approach demonstrates potential generalization capabilities with low data cost in outdoor scenes and another type of event camera.

\paragraph{Limitations.}
As shown in \cref{fig:failure cases}, when overexposure and motion blur issues are observed together with challenging hand poses, it is challenging for \methodname{} to output proper predictions.
Besides, the respective performance from complementary use of event streams and RGB frames in our experiment is affected by the different pixel resolutions.
As event cameras evolve, we expect future work to collect data from event cameras with higher image resolution and lower noise to rigorously validate the effects of complementing event streams and RGB images.

\section*{Acknowledgements}
This work is supported by Beijing Natural Science Foundation (Grant No. L233024, L232028), and National Natural Science Foundation of China (Grand No. 62136001, 62088102).

\newpage
{
    \small
    \bibliographystyle{ieeenat_fullname}
    \bibliography{main}
}

\clearpage
\setcounter{page}{1}
\maketitlesupplementary
\appendix


\paragraph{Overview.}
In the supplemental materials, we first introduce the details of indoor and outdoor real world datasets and synthetic dataset in \cref{supple: sec: datasets}. 
Then we show supplemental experiment results in \cref{supple: sec: experiment results}.
Finally, we illustrate the details of comparison methods in \cref{supple: sec: comparison of methods} and the implementation details in \cref{supple: sec: implementation details}.

\begin{figure}[t]
  \centering
  \includegraphics[width=\linewidth]{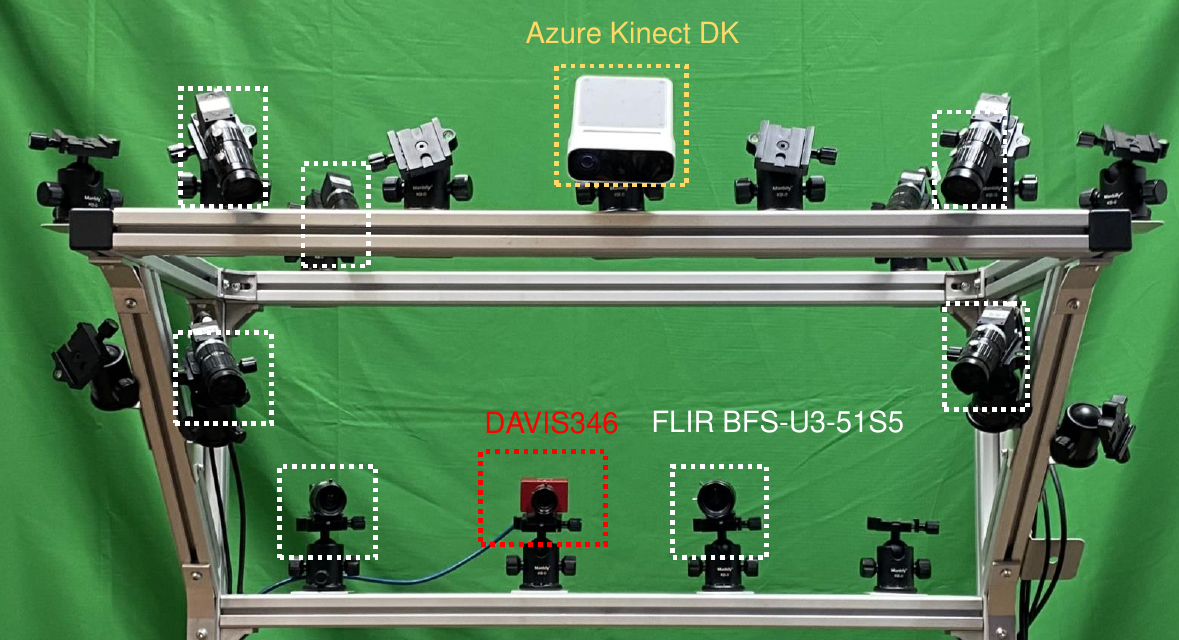}
  \caption{
    Multi-camera system for capturing indoor sequences.
    An event camera (DAVIS346, red circle) is synchronized with 7 RGB cameras (FLIR BFS-U3-51S5, yellow circles) to capture multi-view RGB images and monocular event streams.
    An RGB-D camera (Azure Kinect DK, white circle) is used as an auxiliary camera in the calibration step for precise calibration.
  }
  \label{fig:multi-camera system}
\end{figure}

\begin{figure}[t]
  \centering
  \captionsetup[subfigure]{labelformat=empty}
  \subfloat[\vspace{0.15cm}\hspace{-0.17cm}Normal scenes]{
     \centering
     \hspace{-0.17cm}\includegraphics[width=\linewidth]{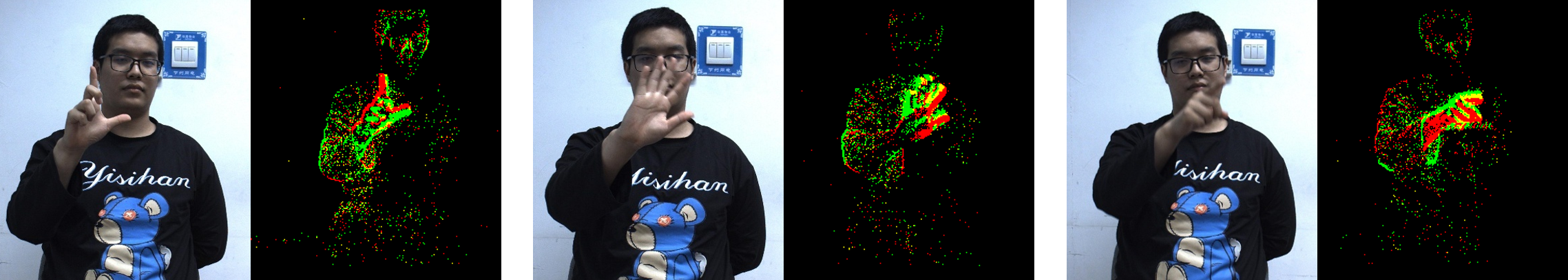}
  }\\
   \subfloat[\vspace{0.15cm}\hspace{-0.17cm}Strong light scenes]{
     \centering
     \hspace{-0.17cm}\includegraphics[width=\linewidth]{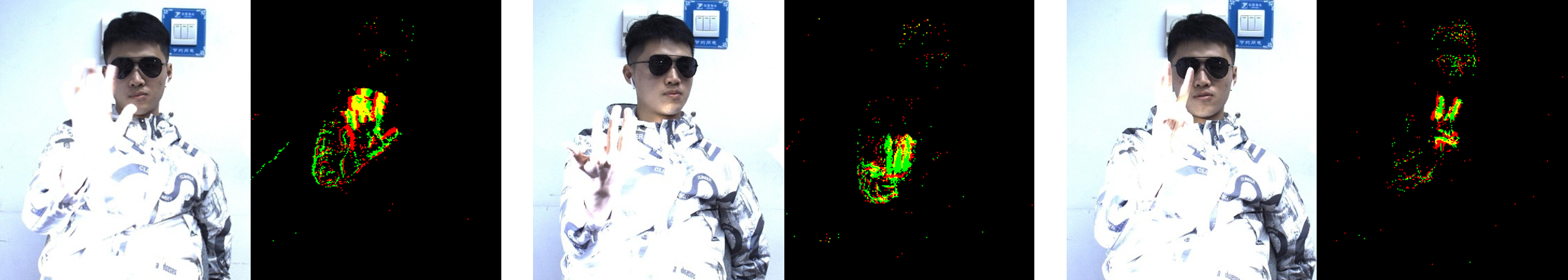}
  }\\
    \subfloat[\vspace{0.15cm}\hspace{-0.17cm}Flash scenes]{
     \centering
     \hspace{-0.17cm}\includegraphics[width=\linewidth]{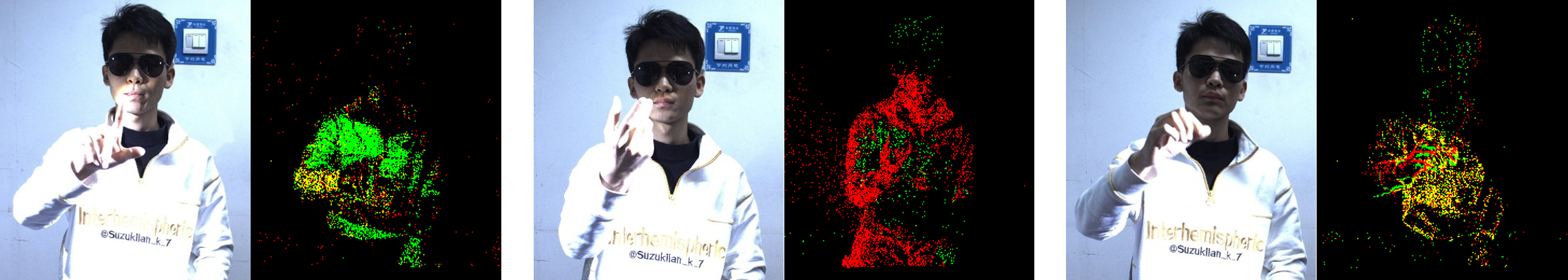}
  }\\
    \subfloat[\vspace{0.15cm}\hspace{-0.17cm}Fast motion scenes]{
     \centering
     \hspace{-0.17cm}\includegraphics[width=\linewidth]{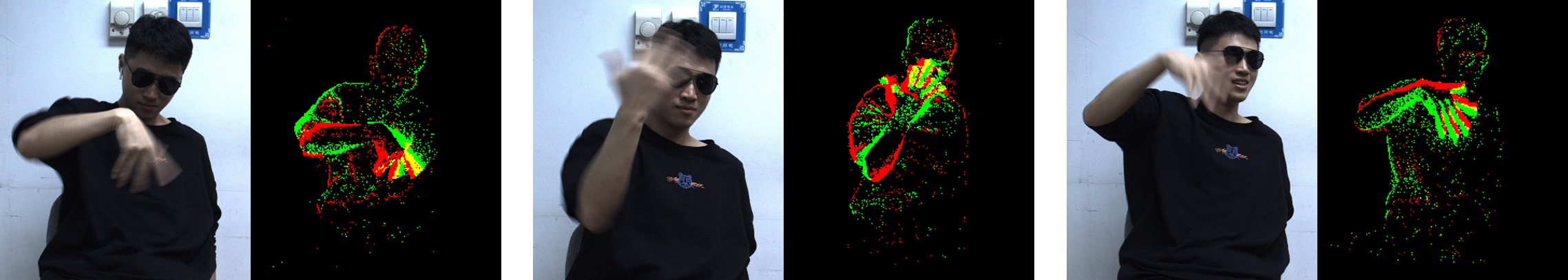}
  }
  \caption{
  Examples of indoor sequences from {\sc EvRealHands}.
  RGB frames (left) and corresponding event streams (right) in normal, strong light, flash and fast motion scenes.
  }
  \label{fig:real data}
\end{figure}

\section{Datasets}
\label{supple: sec: datasets}

To supplement the section of Datasets in the main paper, we show details about the indoor and outdoor sequences of {\sc EvRealHands} and simulation process of the synthetic data. 

\subsection{Indoor Sequences}
\paragraph{Capture system.}
{The indoor sequences of \sc EvRealHands} is captured in a multi-camera system~\cite{handdataset17, megatrack20}.
As shown in \cref{fig:multi-camera system}, in our multi-camera system, 7 RGB cameras (FLIR, 2660$\times$2300 pixels) and an event camera (DAVIS346, 346$\times$260 pixels) capture data from different views simultaneously.
After synchronizing all the cameras with an external 15 Hz Transistor-Transistor Logic (TTL) signal, we calibrate all the cameras with a moving chessboard~\cite{calibration97} with RGB images from FLIR camera, APS frame from DAVIS346, and depth images from the RGB-D camera.

\paragraph{Data acquisition.}
We show examples from our dataset in \cref{fig:real data}.
In the sequence of normal scenes, we capture RGB images without motion blur under everyday indoor lighting.
When subjects keep hands static, the foreground scarcity issue of event-based Hand Mesh Reconstruction (HMR) appears.
We capture 457 seconds of data under strong light by keeping two glare flashlights on with 2000 lumen.
We set the exposure time of 6 annotation RGB cameras to 0.5 ms to avoid overexposure and that of 1 reference RGB camera to 15 ms to make its RGB images overexposed.
Therefore, we obtain images with high-quality from annotation cameras for multi-view annotation and overexposed images from the reference camera as training and evaluation data.
To simulate background overflow issue, we collect sequences under flash light of 317 seconds by making flashlights strobe at 1 Hz. 
Besides, we also collect 69 seconds of fast motion sequences.
To simulate motion blur issues of RGB-based HMR, the subjects shake hands rapidly and fingers appear as ghost in the images.

\paragraph{Annotation.}
Following \cite{interhand20}, we first annotate 21 2D keypoints on each RGB view with Mediapipe~\cite{zhang2020mediapipe} and correct the unqualified annotations manually.
By triangulating 2D keypoints from 7 RGB views, we obtain 3D joints. Then we fit the MANO model to the 3D joints to get the hand shape for each timestamp.

\subsection{Outdoor Sequences.}

\begin{figure}[t]
  \centering
  \includegraphics[width=\linewidth]{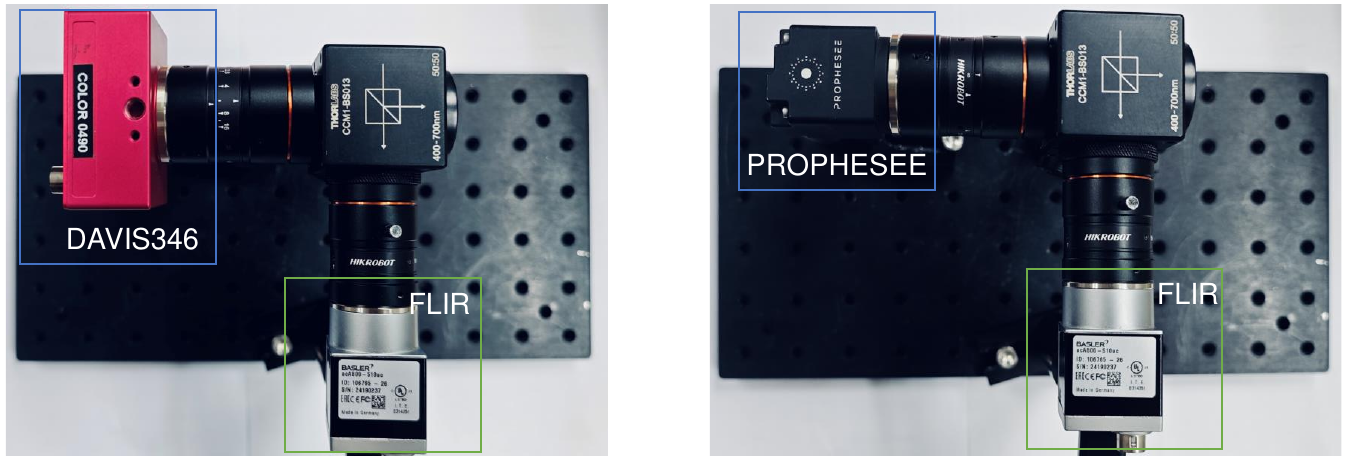}
  \caption{
    Hybrid camera system with an event camera and an RGB camera.
  }
  \label{fig:outdoor-camera system}
\end{figure}

\begin{figure}[t]
  \centering
  \includegraphics[width=\linewidth]{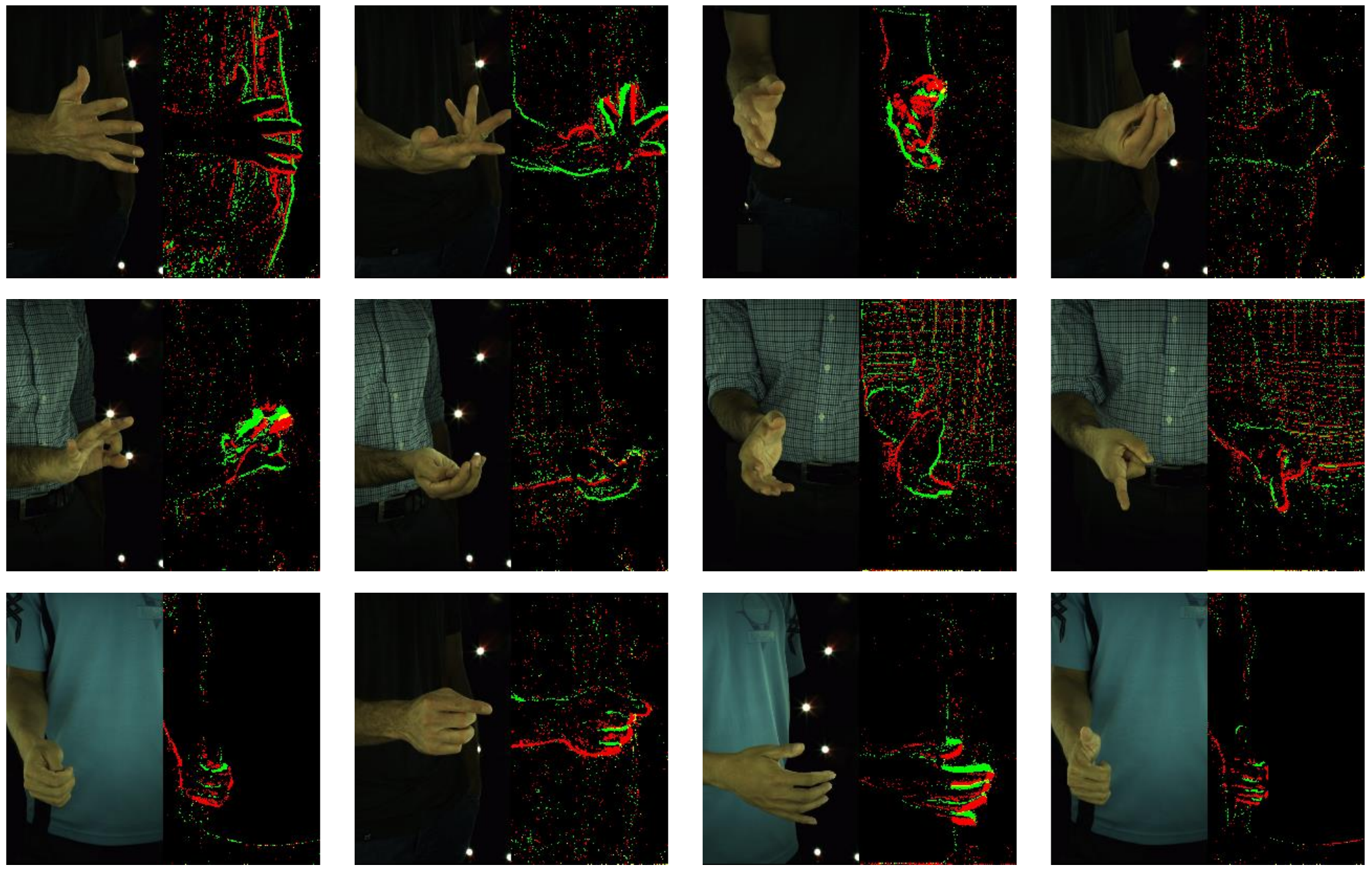}
  \caption{
    Visualization of our synthetic dataset generated using {\sc InterHand2.6M}~\cite{interhand20} and v2e event simulator~\cite{v2e21}.
  Examples of RGB frames (left) and corresponding event streams (right) are displayed side by side.
  }
  \label{fig:synthetic data}
\end{figure}

\paragraph{Capture system.}
In order to collect data for qualitatively evaluation of the generalization performance of existing methods in outdoor scenarios, we build a hybrid camera system to collect data for qualitatively measuring the generalization performance of existing methods in outdoor scenarios.
As shown in \cref{fig:outdoor-camera system}, the hybrid camera system consists of an RGB camera (FLIR BFS-U3-51S5), an event camera (DAVIS346 Mono or PROPHESEE GEN 4.0) and a beam-splitter (Thorlabs CCM1-BS013).

\paragraph{Data acquisition.}
We collected 12 sequences of 240 seconds from three subjects, of which 6 sequences are captured using DAVIS346 and the rest using PROPHESEE. 
The outdoor sequences face challenging issues, such as varying natural light conditions, pedestrian interference, and motion blur (including 6 sequences with fast motion). 

\subsection{Synthetic data}
Although EventHands~\cite{eventhands21} proposes a synthetic dataset to the community, there exists domain gap between the used synthetic pose and real-world pose.
Therefore, we use the event simulator v2e~\cite{v2e21} to synthesize event streams from a large-scale RGB-based sequential hand dataset {\sc Interhand2.6M}~\cite{interhand20}.
{\sc Interhand2.6M} captures 2.6 million images from 80$\sim$140 multi-view cameras with various hand poses.
Considering that the image resolution (512$\times$334 pixels) in {\sc InterHand2.6M} is different from that of DAVIS346 camera, we first use affine transformation to warp the RGB images as the same scale of real-world event streams (346$\times$260 pixels) and feed them into the v2e simulator~\cite{v2e21} to get synthetic event streams.
In our synthesizing setup, the positive threshold is set as 0.143 and the negative threshold is 0.225. RGB frames are interpolated ten times to increase the time resolution of synthetic events.
In our experiment, we select the right hand sequences of 9 camera views from 4 subjects.


\section{Supplemental experiment results}
\label{supple: sec: experiment results}
To further evaluate our proposed method, we will illustrate evaluation metrics in \cref{evaluation metrics}, show additional qualitative results in \cref{supple: exper: qualitative} and introduce more quantitative results in \cref{supple: exper: quantitative}.

\begin{figure*}[p]
    \centering
   \includegraphics[width=\textwidth]{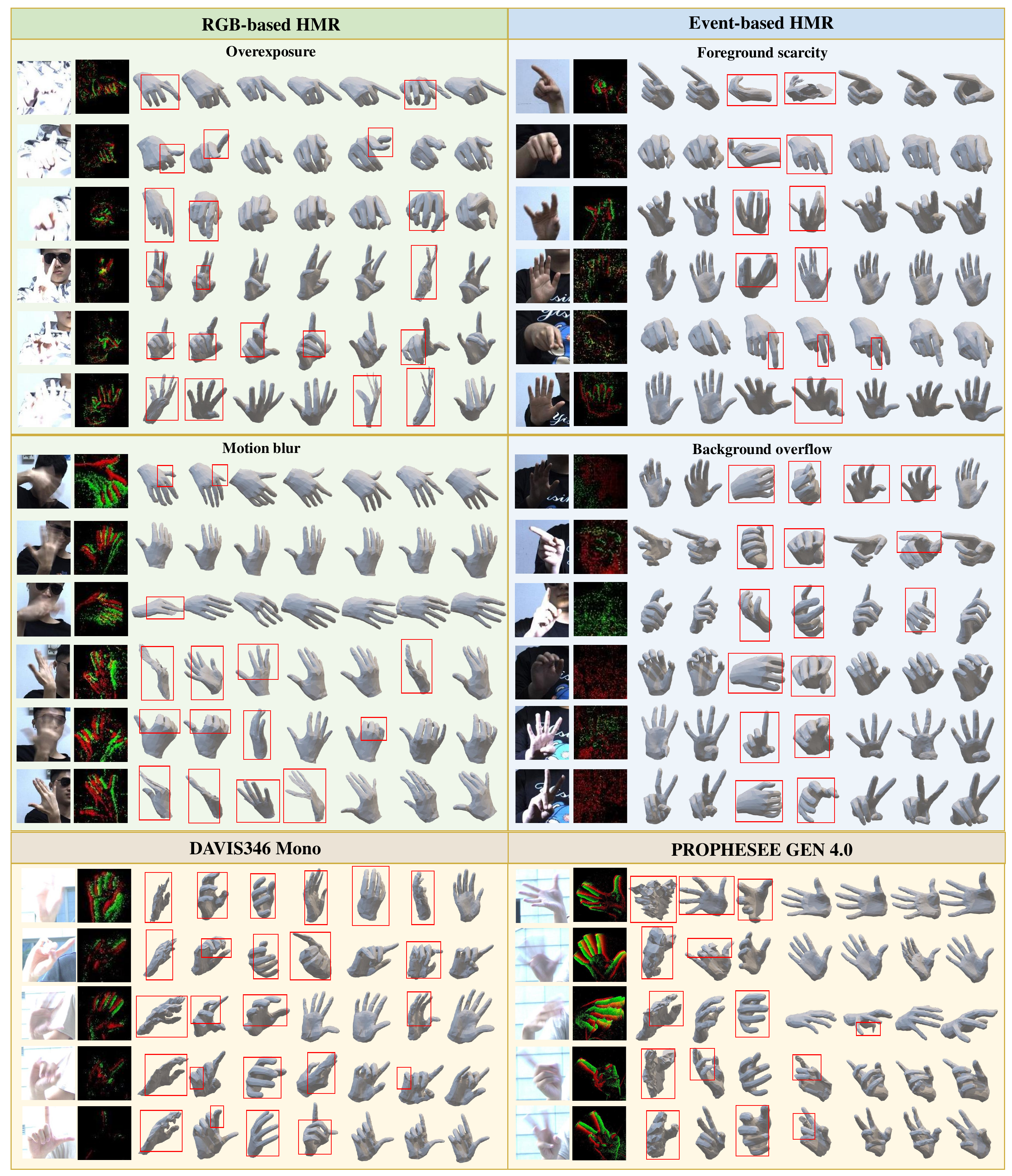}
    \vspace{-0.7cm}
  \begin{flushleft}
    	\scriptsize  \hspace{0.55cm}RGB~ \hspace{0.25cm} Events~ \hspace{0.1cm} MG~\cite{meshgraphormer21} \hspace{0.02cm} FR~\cite{fastmetro22} \hspace{0.05cm} EH~\cite{eventhands21}  \hspace{0.13cm}~ FE\hspace{0.02cm}\quad~ Vanilla \hspace{0.05cm} w/o Deg \hspace{0.1cm} Ours  \qquad~ \hspace{0.05cm} RGB~ \hspace{0.2cm} Events~ \hspace{0.1cm} MG~\cite{meshgraphormer21} \hspace{0.1cm} FR~\cite{fastmetro22} \hspace{0.05cm} EH~\cite{eventhands21}  \hspace{0.2cm}~ FE\hspace{0.03cm}\quad~ Vanilla \hspace{0.15cm} w/o Deg \hspace{0.1cm} Ours
  \end{flushleft}
  \caption{Additional qualitative analysis of HMR methods under challenging issues (green box titled with `\textit{RGB-based HMR}' and blue box titled with `\textit{Event-based HMR}'), outdoor scenes (camel box titled with `\textit{DAVIS346 Mono}' ), and PROPHESEE sequences (camel box titled with `\textit{PROPHESEE GEN 4.0}'). For each issue, columns from left to right are RGB images, events, results from Mesh Graphormer (MG)~\cite{meshgraphormer21}, FastMETRO-RGB (FR)~\cite{fastmetro22}, EventHands (EH)~\cite{eventhands21}, FastMETRO-Event (FE), \methodname{}-vanilla (Vanilla), \methodname{} without \degradation{} (w/o Deg) and \methodname{} (Ours).}
    \label{fig: additional quality results}
\end{figure*}


\begin{figure*}[t]
    \centering
    \subfloat{\includegraphics[width=0.33\textwidth]{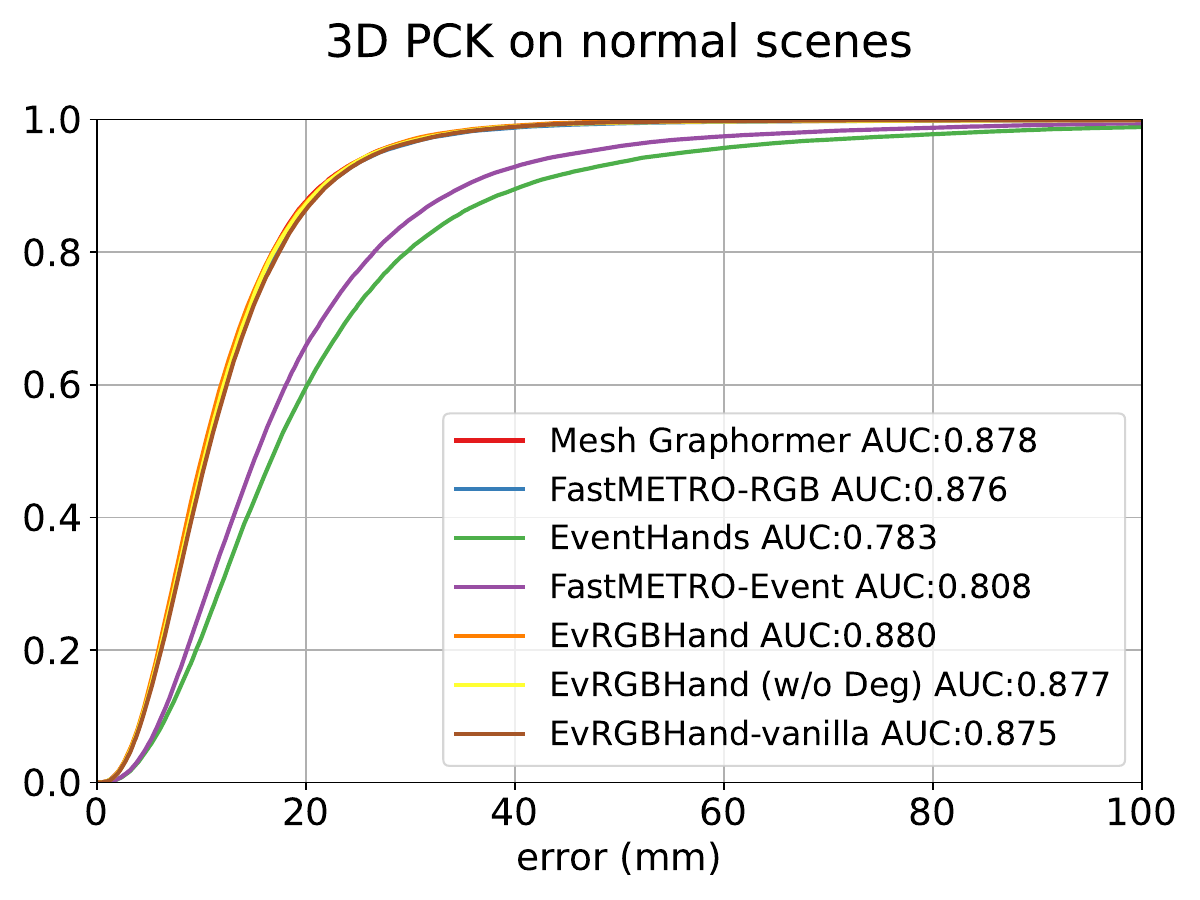}}
    \subfloat{\includegraphics[width=0.33\textwidth]{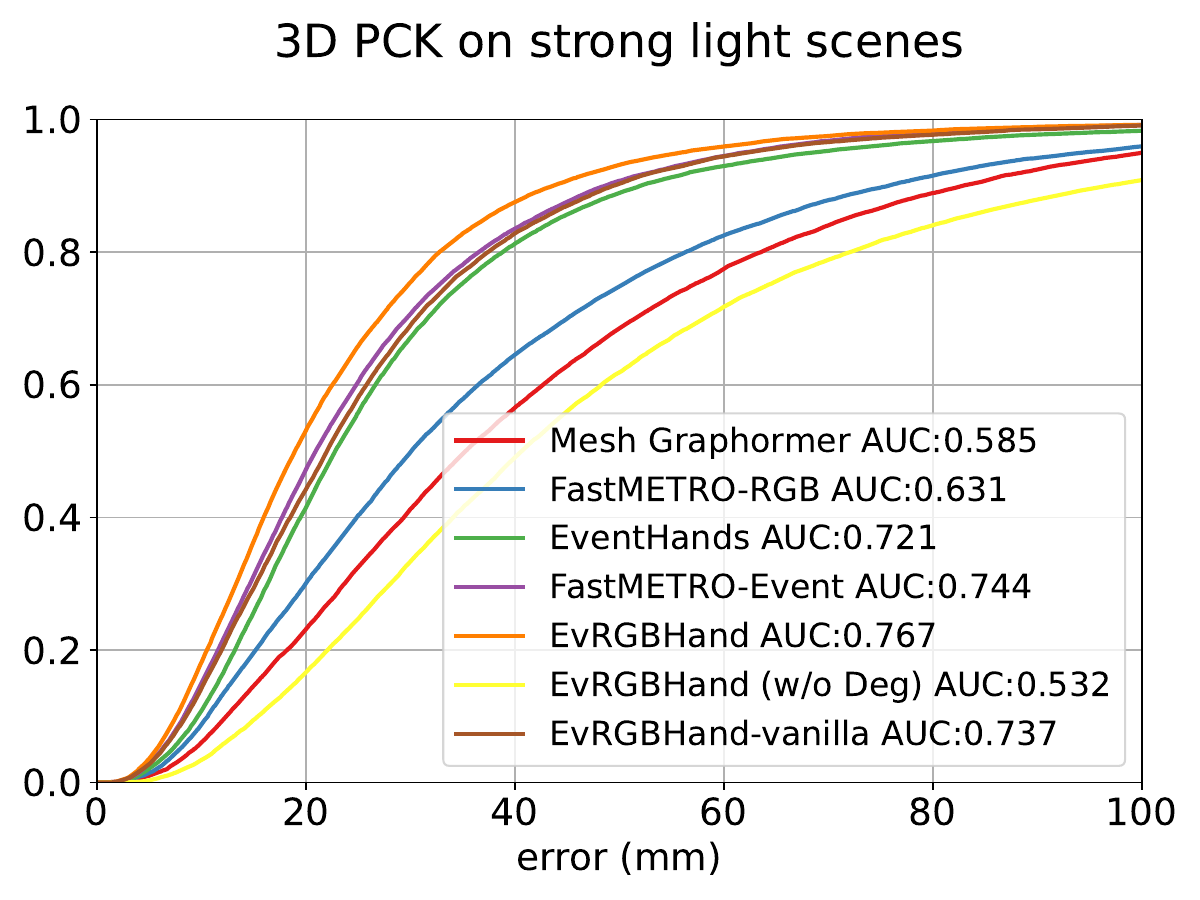}}   
    \subfloat{\includegraphics[width=0.33\textwidth]{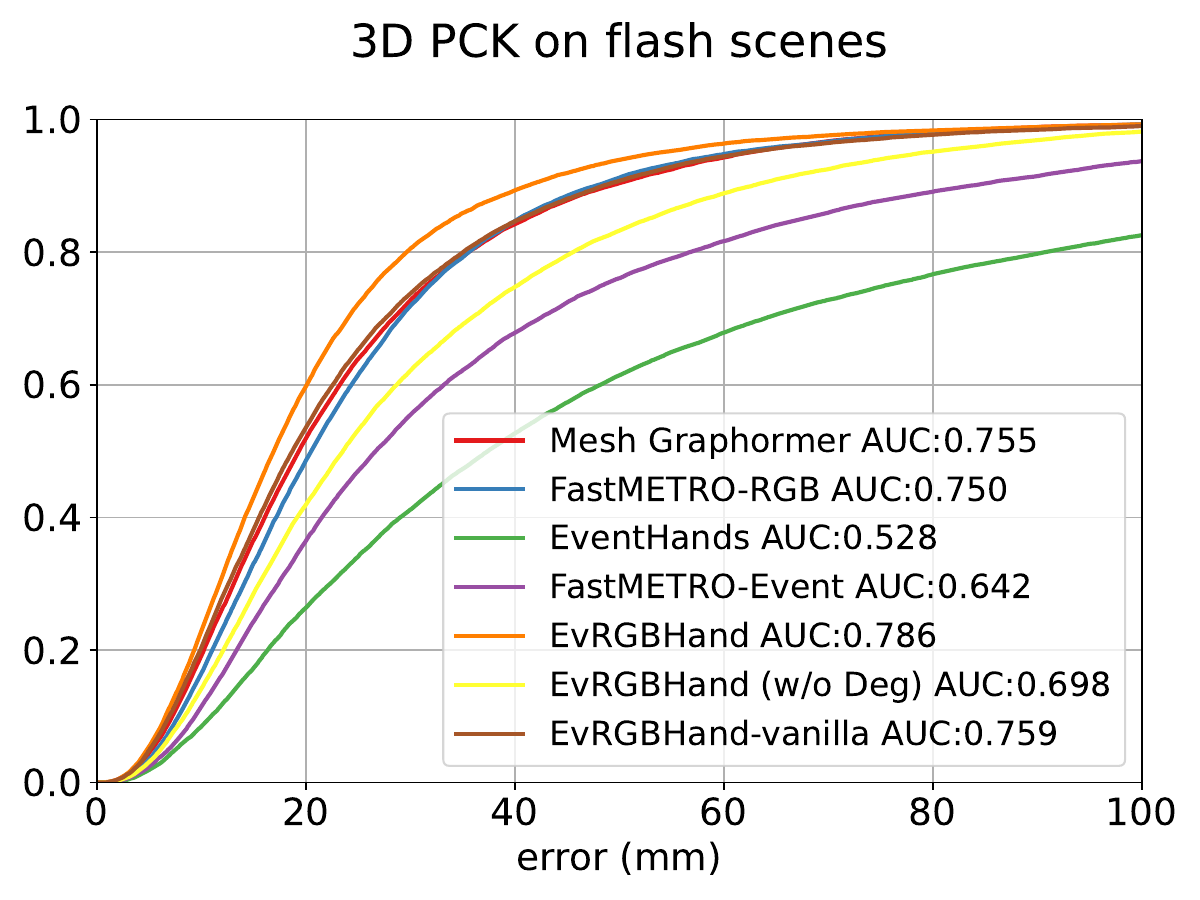}}\\
    \caption{3D PCK curves of \methodname{} and other baselines.}
    \label{fig: AUC curves}
\end{figure*}

\subsection{Evaluation metrics}
\label{evaluation metrics}
\noindent \textbf{Accuracy.}
MPJPE/MPVPE
is root-aligned mean per joint/vertex position error in Euclidean distance (mm).
It measures the distance between predicted and ground truth results. PA-MPJPE/PA-MPVPE measures the MPJPE/MPVPE between ground truth coordinates and 3D aligned predicted coordinates using Procrustes Analysis (PA)~\cite{pa75}.
This metric ignores the scale and global rotation.
AUC is the area under the curve of PCK (percentage of correct keypoints) with thresholds ranging from 0$\sim$100 mm for 3D annotated sequences. 
The lower the metrics above are, the better, except for AUC.

\vspace{2mm}
\noindent \textbf{Computational cost.}
FLOPs is the floating point operations per inference and Params is the count of parameters.

\subsection{More qualitative results}
\label{supple: exper: qualitative}

As shown in \cref{fig: additional quality results}, we show more qualitative results of the comparison between \methodname{} and other baselines.
These qualitative results demonstrate the complementary effects and generalization ability of \methodname{} for HMR with events and images.

To fully leverage the high temporal resolution property of event cameras, we achieve high frame rate inference via an asynchronous fusion strategy. Specifically, the event stream with high temporal resolution can be split into discrete temporal bins. These bins, representing discrete event intervals, are configured to surpass the frame rate of traditional RGB cameras in frequency. Subsequently, each of these temporal bins undergoes fusion with the latest RGB frame, facilitated by EvImHandNet. The temporal relationship between the timestamp $t_i$  of an event bin and the timestamp $t_j$ of the corresponding RGB frame can be formulated as follows:
\begin{equation}
j=\underset{k}{\arg \min }\left|t_i-t_k\right|, t_i - t_k \geq 0.
\end{equation}


    

\subsection{3D PCK curves and AUC.}
\label{supple: exper: quantitative}
We show 3D PCK curves of the baselines and \methodname{} under several scenes in \cref{fig: AUC curves}.
The results show that \methodname{} outperforms all the methods based on a single sensor on AUC.
By complementary usage of events and images, \methodname{} achieves a higher AUC (0.07 $\sim$ 0.14) than event-based HMR on normal scenes and flash scenes, and RGB-based HMR on strong light scenes.

\section{Details of comparison methods}
\label{supple: sec: comparison of methods}

\begin{figure}[t]
  \centering
  \includegraphics[width=\linewidth]{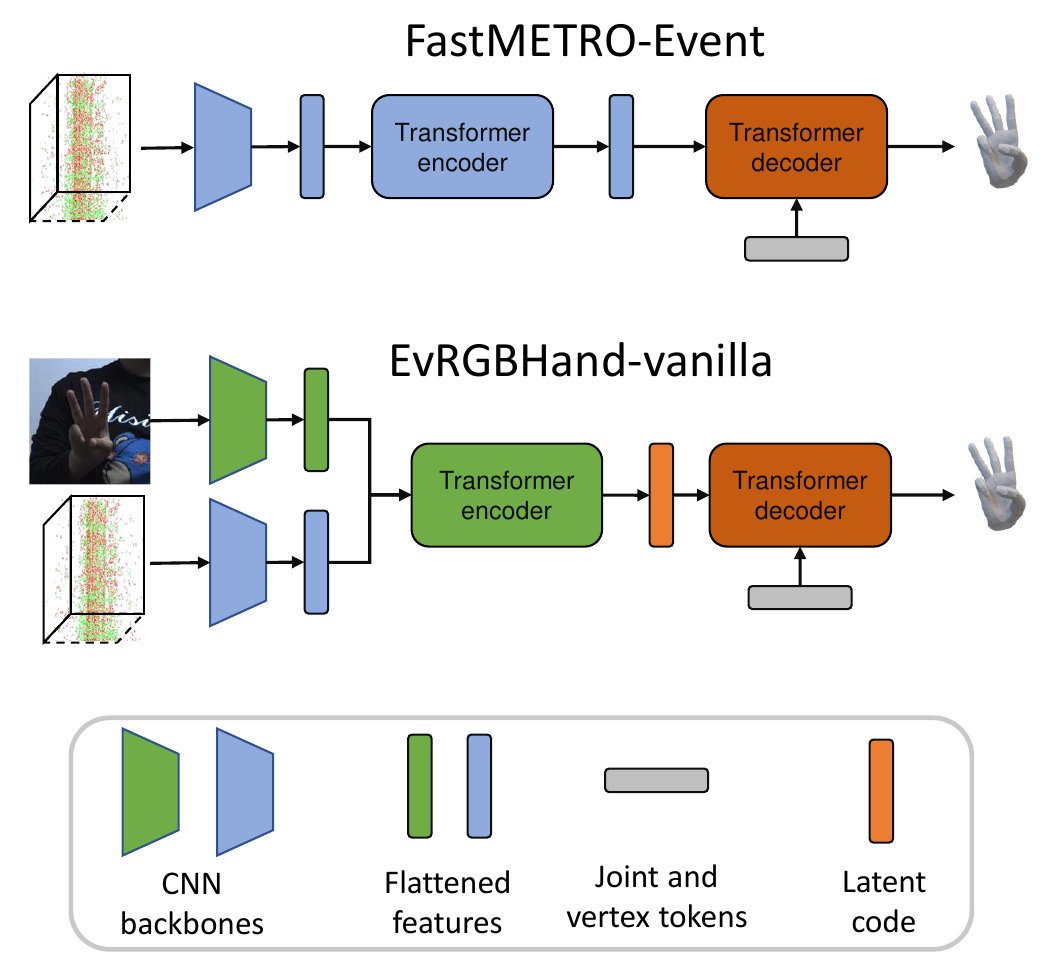}
  \caption{
    Brief structures of FastMETRO-Event and \methodname{}-vanilla proposed in the main paper.
  }
  \label{fig:method comparison}
\end{figure}

As shown in \cref{fig:method comparison}, we provide additional explanations about the structures of FastMETRO-Event and \methodname{}-vanilla. 
FastMETRO-Event derives from the RGB-based HMR approach, FastMETRO~\cite{fastmetro22}. 
FastMETRO~\cite{fastmetro22} is an encoder-decoder based transformer framework by disentangling the image embedding and mesh estimation, which can achieve fast convergence, low computation cost, and comparable accuracy.
The only difference between FastMETRO-Event and FastMETRO~\cite{fastmetro22} lies in the input: FastMETRO-Event utilizes an event representation instead of an RGB image. 
Despite this simple substitution, it has outperformed the current state-of-the-art event-based method, EventHands~\cite{eventhands21}.

\methodname{}-vanilla is built upon the FastMETRO~\cite{fastmetro22} framework, integrating event features and image features as tokens into a transformer encoder. This approach follows the fashion of contemporary multi-modal fusion methods~\cite{flava22, perceiver21, VATT21}.

\section{Implementation details}
\label{supple: sec: implementation details}

For event representation, we set $N=7000$ for evaluation. While for training step, the number of events in each stacked event frame is selected randomly from 5000 $\sim$ 9000 for data augmentation.
We apply geometric augmentation including scale, rotation and translation.

The details of \degradation{} are as follow:
\begin{itemize}
    \item \textbf{Overexposure (OE):} Color jitter augmentation is adopted with a probability of 0.4 to change the image brightness. And the brightness factor is randomly selected from 0.8 to 4.
    \item \textbf{Motion blur (MB):} Motion blur augmentation is applied with a 0.3 probability. To synthesize blurry images, we first apply video interpolation via estimated optical flow to increase 15 fps videos to 120 fps ones. Then a single blurry hand image is generated by averaging 17 consecutive frames, which are interpolated from 3 sharp sequential frames.
    \item \textbf{Background overflow (BO):} 
    Salt-and-pepper noise is applied to each pixel with a probability of 0.2.
\end{itemize}
Moreover, event camera will emit temporally noisy outputs caused by the quantal nature of photons and events with leak noise from junction leakage and parasitic photocurrent~\cite{noise17, v2e21}.
These noises are noticeable in strong light and flash scenes.
For data augmentation on event streams, we add Gaussian noise with a probability of 0.8 on event streams to simulate temporal noise.
The deviation of Gaussian noise is randomly selected from 0.05 to 0.2.

In order to effectively extract hand features, we crop the frames with bounding boxes.
We first obtain 3D joints at the target time by linear interpolation (specially for the stacked event frame) and project the 3D joints onto the image plane to get 2D keypoints, which can be exactly covered by an rectangle.
The bounding box is a square which shares the same center with the rectangle and has 1.6 times the length of the longer side of the rectangle.
The sizes of bounding boxes are 192$\times$192 for both RGB frames and stacked event frames.
In our experiments, we use ResNet~\cite{resnet16} as our CNN backbones.
The number of transformer blocks $L$ is set to 3 and the hidden state dimensions of $L$ blocks are 256.
The number of transformer heads is set to 8.
For the vertex and joint loss functions, $\lambda_{\mathbf{V}}$ is 100 and $\lambda_{\mathbf{J}}$ is 2000.
The initial learning rate is set to 0.0001 and we apply a cosine annealing schedule~\cite{coslr17}.
We use Adam~\cite{adam15} as the optimizer with $\beta_{1}=0.9$, $\beta_{2}=0.999$ and no weight decay.
We train \methodname{} with a batch size of 32 for 50 K iteractions on 2 NVIDIA TITAN X GPUs.



\end{document}